\DocumentMetadata{
      pdfversion=2.0,pdfstandard=ua-2,
      testphase={phase-III,firstaid,math,title}
    }

\documentclass[sigconf,screen,nonacm]{acmart-tagged}
\AtBeginDocument{%
  }

\setcopyright{none}
\setcctype{by}
\acmDOI{10.1145/3757279.3785549}
\acmYear{2026}
\copyrightyear{2026}
\acmISBN{979-8-4007-2128-1/2026/03}
\acmConference[HRI '26]{Proceedings of the 21st ACM/IEEE International Conference on Human-Robot Interaction}{March 16--19, 2026}{Edinburgh, Scotland, UK}
\acmBooktitle{Proceedings of the 21st ACM/IEEE International Conference on Human-Robot Interaction (HRI '26), March 16--19, 2026, Edinburgh, Scotland, UK}
\received{2025-09-30}
\received[accepted]{2025-12-01}

\usepackage{balance}
\usepackage{multirow}
\usepackage{mathtools}
\usepackage{stmaryrd}
\usepackage{listings}
\usepackage{xcolor}
\usepackage{float}
\usepackage{dirtytalk}
\newcommand{\hired}[1]{{\color{magenta} \textit{\textbf{#1}}}}

\newcommand{\higreen}[1]{{\color{teal} \textit{\textbf{#1}}}}
\newcommand{\hiblue}[1]{{\color{blue} \textit{\textbf{#1}}}}
\usepackage{microtype}

\newcommand{\interp}{\textsc{InterPReT}}

\lstdefinestyle{mypython}{
    language=Python,
    basicstyle=\ttfamily\footnotesize\fontfamily{cmtt}\selectfont,
    keywordstyle=\color{blue},
    stringstyle=\color{red},
    commentstyle=\color{green!50!black},
    showstringspaces=false,
    numberstyle=\tiny\color{gray},
    numbers=left,
    stepnumber=1,
    breaklines=true,
    escapeinside={||},
    frame=single,
    texcl=true,
}

\begin{document}

\title{InterPReT: Interactive Policy Restructuring and Training Enable Effective Imitation Learning from Laypersons}

% \affiliation{%

% \author{Feiyu Gavin Zhu}
% \affiliation{%
%   \institution{Carnegie Mellon University}
%   \city{Pittsburgh}
% \email{feiyuz@andrew.cmu.edu}
% \author{Jean Oh}
% \affiliation{%
%   \institution{Carnegie Mellon University}
%   \city{Pittsburgh}
% \email{jeanoh@cmu.edu}
% \author{Reid Simmons}
% \affiliation{%
%   \institution{Carnegie Mellon University}
%   \city{Pittsburgh}
% \email{rsimmons@andrew.cmu.edu}

\author{Feiyu Gavin Zhu}
\email{feiyuz@andrew.cmu.edu}
\affiliation{%
  \institution{Carnegie Mellon University}
  \city{Pittsburgh}
  \state{Pennsylvania}
  \country{USA}
}

\author{Jean Oh}
\email{jeanoh@cmu.edu}
\affiliation{%
  \institution{Carnegie Mellon University}
  \city{Pittsburgh}
  \state{Pennsylvania}
  \country{USA}
}

\author{Reid Simmons}
\email{rsimmons@andrew.cmu.edu}
\affiliation{%
  \institution{Carnegie Mellon University}
  \city{Pittsburgh}
  \state{Pennsylvania}
  \country{USA}
}

\renewcommand{\shortauthors}{Feiyu Gavin Zhu, Jean Oh, and Reid Simmons}

\begin{abstract}

Imitation learning has shown success in many tasks by learning from expert demonstrations.
However, most existing work relies on large-scale demonstrations from technical professionals and close monitoring of the training process.
These are challenging for a layperson when they want to teach the agent new skills.
To lower the barrier of teaching AI agents, we propose \underline{Inter}active \underline{P}olicy \underline{Re}structuring and \underline{T}raining (\interp{}), which takes user instructions to continually update the policy structure and optimize its parameters to fit user demonstrations.
This enables end-users to interactively give instructions and demonstrations, monitor the agent's performance, and review the agent's decision-making strategies.
A user study ($N=34$) on teaching an AI agent to drive in a racing game confirms that our approach yields more robust policies without impairing system usability, compared to a generic imitation learning baseline, when a layperson is responsible for both giving demonstrations and determining when to stop.
This shows that our method is more suitable for end-users without much technical background in machine learning to train a dependable policy.\footnote{This is a preprint version. The final version will appear in the Proceedings of the 21st ACM/IEEE International Conference on Human-Robot Interaction. Project page at \href{https://zfy0314.github.io/interpret/}{https://zfy0314.github.io/interpret/}.}

\end{abstract}





\maketitle

\section{Introduction}

As each individual has their own preferences, needs, and living spaces, robots need to be able to easily acquire new skills from their users to fit their specific requirements and environments.
Despite advances in imitation learning policy representations \cite{zhao2023learning, chi2024diffusionpolicy}, more intuitive data collection hardware \cite{chi2024universal, liu2025factr}, and better human-robot joint learning paradigms \cite{luo2025human}, whether a layperson can take advantage of them remains largely underexplored. 

\begin{figure}[t]
    \centering
    \includegraphics[width=0.7\columnwidth]{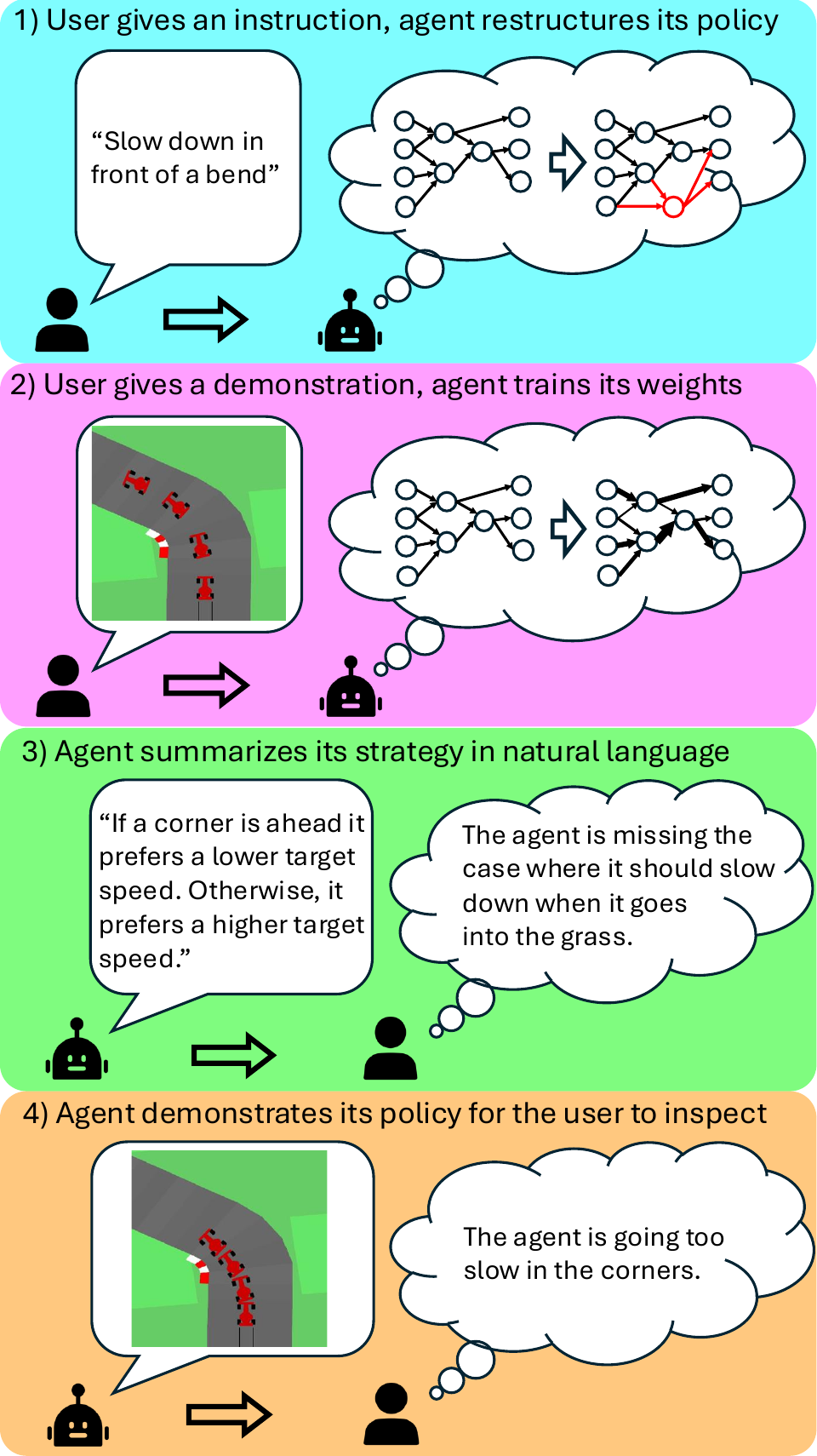}
    \caption{
        Interaction modes in \interp{}.
        % There are four modes of interaction:
        % 1) Human teacher gives an instruction and the agent restructures its policy representation; 
        % 2) Human teacher gives a demonstration and the agent trains the weights in its policy;
        % 3) Agent reflects its current strategy and the human teacher updates their instructions;
        % 4) Agent demonstrates its policy and the human teacher knows what to teach next time.
        The user repeatedly interacts with the agent until they are satisfied. % with its performance.
    }
    \Description{
        A figure of the four models of interaction between the user and the agent in a two-by-two grid layout.
        The upper-left diagram titles "User gives an instruction, agent restructures its policy".
        It show a human icon saying "Slow down in front of a bend" in a text bubble with an arrow pointing to a robot icon on the right.
        The robot icon shows a thinking cloud bubble that depicts its restructuring process by updating from a graph structure with black nodes and edges into another graph structure with a new node highlighted in red.
        The lower-left diagram titles "User gives a demonstration, agent trains its weights.
        It shows a human icon giving a demonstration in a text bubble showing a car racing task where a red race car is turning left at a corner on a gray race track surrounded by green grass.
        There are four red race cars shown on the track in a line showing the race car at four different time steps.
        They are evenly spaced with a gap width of approximately one car length.
        There is an arrow pointing to the robot icon on the right.
        The robot icon shows a thinking cloud bubble that depicts its learning process by updating from a graph structure with black nodes and edges into another graph structure with different edge widths.
        The upper-right diagram titles "Agent summarizes its strategy in natural langauge".
        It shows a robot icon saying "If a corner is ahead it prefers a lower target speed. Otherwise, it prefers a higher target speed" in a text bubble with an arrow point to a human icon on the right.
        The human icon has a cloud bubble that says "The agent is missing the case where it should slow down when it goes into the grass."
        The lower-right diagram titles "Agent demonstrations its policy for the user to inspect.
        It shows a robot icon giving a demonstration on a gray race track surrounded by green grass turning left at a corner marked with corner markers on the left.
        There are four red race cars shown on the track in a line showing the race car at four different time steps.
        The are cluttered together with each race car almost touching the other race car in front and behind it. 
        There is an arrow between the robot icon and a human icon on the right.
        The human icon says "The agent is going too slow in the corners" in a cloud bubble.
    }
    \label{fig:overview}
\end{figure}

Learning from laypeople differs from learning from technical experts.
The general public does not necessarily know which kinds of demonstration are better suited for imitation learning \cite{saxena2025matters} or when the learned policy is good enough to stop training \cite{hussenot2021hyperparameter}.
In addition to the lack of machine learning knowledge, simply using an unfamiliar teleoperation device can slow a layperson by 4-7x when performing a daily task \cite{wu2024gello}, and collecting a large amount of data can be cognitively tiring \cite{shao2019mental}.
When evaluating general imitation learning algorithms, existing work often provides human demonstrators with very detailed instructions and trains them extensively before using their data to train the policy \cite{zhao2024aloha, hu2025rac} or relies on technical members of the research team \cite{o2024open}.
Existing work oriented to laypeople focuses primarily on accounting for the imperfections in human teachers \cite{kessler2021interactive, cao2021learning}, or application to a particular group of users (e.g., people with motor impairment) \cite{nanavati2025lessons}.

To make general imitation learning more user-friendly, some existing work has explored learning from a few demonstrations with the aid of language instructions \cite{zhu2025sample}, aiming at reducing the number of demonstrations needed by creating a more robust policy representation.
However, it still requires the human expert to specify all the domain knowledge in detail in one go, which is very challenging for most laypeople when the task is non-trivial \cite{babe2024studenteval}.

To solve this limitation and enable the general public to train policies, we propose an interactive learning paradigm: \underline{Inter}active \underline{P}olicy \underline{Re}structuring and \underline{T}raining (\interp{}).
Specifically, the learning agent maintains a list of instructions and demonstrations, uses the instructions to update its policy structure by leveraging the code synthesis capability of a large language model (LLM), and updates its learnable weights using the demonstrations with the imitation learning objective.
Having an instruction-instantiated policy structure has two distinct advantages for learning from laypeople: 1) the sparsity of the structure enables strategic interpretation of the demonstrations, enforcing the policy only focus on important features and not the imperfections, 
and 2) since the policy structure has semantic meaning, an LLM can translate this structure into natural language and explain the policy strategy to the user as feedback.
A between-subject user study ($N=34$) on non-technical users teaching a driving policy shows that our approach significantly outperforms a generic baseline imitation learning paradigm.

Our contribution is twofold: 1) we introduce a teaching framework that builds a policy from both demonstrations and instructions from end-users through multiturn interactions, and 2) we conducted a user study specifically for learning from laypeople to analyze our framework's ability to produce robust policies, efficiently improve during each round of interaction, and overall usability.

\section{Related Work}

\subsection{Interactive Learning from Human Feedback}

Learning from Demonstration (LfD) aims to learn a policy by imitating expert actions \cite{pomerleau1988alvinn}.
The three main challenges are 1) the distribution shift due to error accumulation \cite{osa2018algorithmic}, 2) causal confusion where the policy is conditioned on task-irrelevant features \cite{de2019causal}, and 3) imperfect demonstrations from the demonstrators \cite{wu2019imitation}.

Previous attempts have explored an interactive teaching paradigm in which the human teacher iteratively refines the learned policy. 
This includes trajectory relabeling \cite{ross2011reduction} with shared autonomy \cite{luo2025human} to reduce human input, imperfection-aware learning algorithms that detect and adjust when the human teacher is not paying attention or giving incorrect feedback \cite{kessler2021interactive, faulkner2023using}, and incorporating various types of feedback \cite{fitzgerald2022inquire} for more efficient human queries.
Others have replaced the human teacher with an LLM \cite{wang2024rl, chen2024elemental}.

% Previous attempts to alleviate distribution mismatch include trajectory relabeling \cite{ross2011reduction} with shared autonomy \cite{luo2025human} and increasing the diversity of demonstrations through domain randomization \cite{tobin2017domain}, data augmentation \cite{zhou2023nerf}, and collecting error recovery data \cite{laskey2017dart}.
% Some work has looked at explicitly modeling the causal structure of the task environment \cite{huang2023went}.
% Others have proposed self-supervised demonstration scoring \cite{cao2021learning} and incorporating preference signals \cite{cao2024limited}.

% Our work also falls under interactive learning from human teachers. 
% In addition to learning from the demonstrations, we also allow the teacher to give instructions to help the agent interpret their demonstrations.
Our work is the most similar to \cite{peng2024pragmatic} where we used LLMs to incorporate semantic information.
But ours allows the instruction-guided policy structure generation in addition to feature engineering.
Compared to previous work on structured policies \cite{zhu2025sample}, we formulate the agent learning process as an interactive process, where demonstrators do not have to get everything right the first time and can continue to inspect what the agent has learned.

\subsection{Complementing Learning with Language}

Given recent advancements in language encoders \cite{wei2022emergent} and multimodal alignment \cite{driess2023palm}, much work has explored how language can be used to augment learning in continuous embodied domains.

Several works have looked at aligning language with task specifications to leverage similarities between tasks \cite{lynch2021languageconditionedimitationlearning, yu2022using}, failure trajectory to learn error recovery \cite{dai2025racer}, and action primitives to take user corrections on the fly \cite{shi2024yell}.
Other works involve expressing actions in language and exploiting the zero-shot capabilities of LLMs to output low-level commands in language-like coding \cite{tang2023saytap}.
Many recent works explored the use of LLMs' coding capability to convert language commands into code for reward design in learning \cite{ma2023eureka, li2025R}, robot plans \cite{liang2022code, zhu2023ghost}, and policy structures \cite{zhu2025sample}.

Our work also follows the language-to-structure generation paradigm for a more efficient interpretation of the demonstrations.
However, instead of relying on long and specific domain knowledge from experts or environment definition, it is designed to take instructions from laypeople who may have general knowledge for the task but could not fully specify all aspects of the policy.

\subsection{Interactive Learning with Code Generation}

% To convert the users' needs to robot policies, many works have investigated how to make it easier for laypeople to program AI agents.
To convert the users' needs to robot policies, many works have investigated how to make programming AI agents more accessible.
Some work has designed interfaces for laypeople to specify an RL problem by defining goal conditions using graphical programming \cite{zhao2022supporting, ayalew2025enabling}, designing finite state machines from instructions \cite{gan2024can}, and injecting constraints into tabular-based policies \cite{brawer2023interactive}.
Others utilize program synthesis on a predefined syntax for web application automation \cite{dong2022webrobot} or long-horizon robot planning \cite{laird2017interactive,patton2024programming}.

% However, most previous work operates in discrete domains and assumes access to a set of predefined function APIs.
Ours is the most similar to \cite{kim2025interactiveprogramsynthesismodeling}, where the policy is neuro-symbolic with the symbolic part generated from instructions and specific numeric parameters learned from demonstrations.
We use the PyTorch library and allow the synthesized program to be any differentiable function.
This gives the user more flexibility and thus can be applied not only to high-level action primitives but also to low-level, high-frequency continuous control tasks.

\begin{figure*}[t]
    \centering
    \includegraphics[height=0.8in]{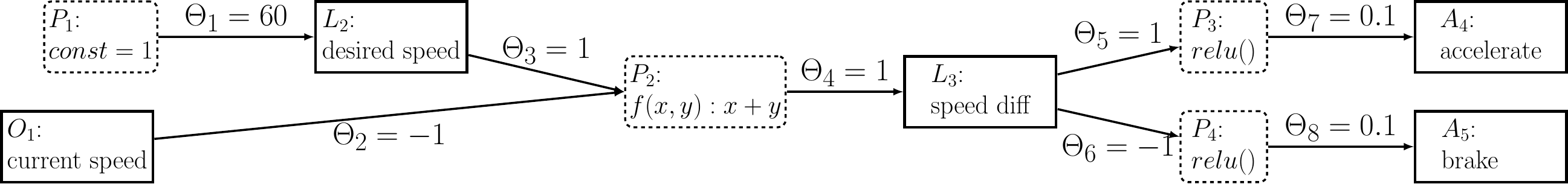}
    \caption{
        A minimal example of a structured policy representing a proportional controller that maintains a constant ``desired speed".
        Solid boxes are variables in $V$ (marked with observation $O$, latent $L$, or action $A$) and dashed boxes are operators in $P$.
        Weights $\Theta$ are associated with the edges.
        During inference, if the observed ``current speed" is $O_1 = 40$, then we propogate the values $P_1 = 1, L_2 = 60, P_2 = 20, L_3 = 20, P_3 = 20, P4 = 0, A_4 = 0.2, A_5 = 0$, and the output action is $\hat a = [0.2, 0]^T$.
    }
    \Description{
        It shows a directed acyclic graph with square nodes connected by edges with weights annotated next to the edges.
        There are 9 nodes in total, organized in 6 columns.
        The left most column has two nodes: the upper one is a dashed square and says P1: const equals 1, the lower one is a solid square that says O1: current speed.
        The second column has one solid square node that says L2: desired speed.
        There is an edge connecting P1 and L2 with Theta1 = 60 annotated on the top.
        The third column has one dashed node that says P2: f(x, y): x + y.
        There is an edge connecting L2 and P2 with Theta3 = 1 annotated on the top.
        There is an edge connecting O1 and P2 with Theta2 = -1 annotated on the bottom.
        There is one solid node on the fourth column that says L3: speed diff.
        There is an edge connected P2 and L3  with Theta4 = 1 annotated on the top.
        there are two dashed nodes on the fifth column: the upper one says P3: relu and the lower one says P4 relu.
        There is an edge connecting L3 and P3 with Theta5 = 1 annotated on the top.
        There is an edge connecting L3 and P4 with Theta6 = -1 annotated on the bottom.
        there are two solid nodes on the sixth column: the upper one says A7: accelerate and the lower one says A5: break.
        There is an edge connecting P3 and A4 with Theta7 = 0.1 annotated on the top.
        There is an edge connecting P4 and A5 with Theta8 = 0.1 annotated on the top.
    }
    \label{fig:policy-structure}
\end{figure*}

\section{InterPReT: Interactive Learning from Both Demonstrations and Instructions}

\subsection{Preliminary: Structured Policy}

We build on previous work on structured policy \cite{zhu2025sample}.
To reduce computation cost and keep the latency low, we use a variant that only uses differentiable operators, such that we do not need to perform grid search on the non-gradient parameters during training.

Mathematically, a structured policy is a 4-tuple $\langle V, P, E, \Theta\rangle$ representing a weighted directed acyclic bipartite graph where
\begin{itemize}
    \item $V$ is a set of nodes that each represents a feature (observed, latent, or action) that is part of the policy decision-making process. They have semantic meanings and represent concepts (e.g., ``target speed").
    \item $P$ is a set of nodes that represents differentiable operations instances (e.g., linear combination or a constant).
    \item $E: \{ \langle u_i, v_i \rangle \in V \times P \cup P \times V \}$ is a set of edges that represents how variables are causally connected through operators.
    % dependencies between latent variables and the operations used to connect them.
    \item $\Theta: E \to \mathbb{R}$ is a set of weights associated with each edge. % They represent the connections between features (e.g., one feature has a causal effect on another). 
\end{itemize}
Typically, the structure is sparse (i.e., $|E|$ is linear to $|V|$) since only variables that are causally related will be connected.
This takes advantage of the sparsity in many tasks \cite{mao2023pdsketch}, has fewer parameters to optimize, and is less susceptible to causal confusion \cite{de2019causal}.
An example structure for maintaining a constant speed is shown in Figure \ref{fig:policy-structure}.
Note that both the observation space $O$ and the action space $A$ are subsets of $V$.
% And $\langle V \cup P, E \rangle$ is a directed acyclic graph.
To simplify the notation, we use $\llbracket\alpha\rrbracket \vcentcolon= \langle V, P, E \rangle$ to denote the structure of the policy.

During inference time, we first assign the values of the observed features in the observation $o$, then iteratively compute the value assignment of each feature node and operator along the edges.
Specifically, the out (head) value of each edge $e = (x, y) \in E$ is
\begin{equation}
e_\text{out} = \Theta_e \cdot x_\text{out}
\end{equation}
Each non-constant operator $p \in P$ takes in the values of the edges that go into it and applies its operation on them:
\begin{equation}
p_\text{out} = p({e_1}_{out}, {e_2}_{out}, \cdots) \text{, where } e_i \in \{ (x, y) \in E \mid y = p \}
\end{equation}
Each non-observed feature $v \in V \setminus O$ has exactly one edge that goes into it $e_v$ and its value is the same as that edge's head value
\begin{equation}
v_\text{out} = {e_v}_\text{out}
\end{equation}
When all feature nodes in $A$ have an assigned value, they will be reconstructed into the action vector $\hat a$. 
We denote this process as
\begin{equation}
    \hat a \vcentcolon= \pi_{\langle \llbracket\alpha\rrbracket, \Theta\rangle}(s)
\end{equation}

In practice, a structured policy is implemented as a PyTorch model \cite{paszke2019pytorch} where the compute graph represents the structure and parameter updates through gradient descent are easily achievable.

\subsection{Agent}

% One major limitation of the previous work is that the generation of policy structures requires a long and exhaustive prompt, which is hard to acquire from laypeople in one attempt.
% Therefore, we allow users to update instructions through interactions.

Formally, each \textit{agent instance} is a $4$-tuple $\mathcal{A}_i = \langle \mathcal{I}_i, \mathcal{D}_i, \llbracket \alpha \rrbracket_i, \Theta_i\rangle$, where $\mathcal{I}$ is the current set of instructions (e.g., ``slow down before turns"), $\mathcal{D}$ is the current set of demonstrations, and $\llbracket \alpha \rrbracket_i$ is as defined above for the policy structure.

\subsection{Policy Training}

Given a structure $\llbracket \alpha \rrbracket_i$ and a set of demonstrations $\mathcal{D}_i$, we can train the weights using the standard imitation learning objective \cite{pomerleau1988alvinn}:

\begin{equation}
    \hat\Theta_i = \text{Train}(\llbracket \alpha \rrbracket_i, \mathcal{D}_i)
    \vcentcolon=\arg\min_\Theta \sum_{s, a \in \mathcal{D}_i} || a - \pi_{\langle \llbracket\alpha\rrbracket_i, \Theta\rangle}(s) ||
\end{equation}
% In practice, since each learnable weight connects two features with semantic meanings, we query an LLM\footnote{All experiments are done with \texttt{gpt-5-mini-2025-08-07} with medium reasoning effort and low verbosity.} to set the initial weights $\Theta_0$ and perform gradient descent with the Adam \cite{kingma2014adam} optimizer. 
In practice, since features and operators have semantic meanings, we query an LLM\footnote{All experiments are done with \texttt{gpt-5-mini-2025-08-07} with medium reasoning.} to estimate the causal effects between them and use it to initialize the weights $\Theta_0$, then perform gradient descent with the Adam \cite{kingma2014adam} optimizer. 
Previous work has shown that having a set of good initial weights stabilizes training and reduces the chance of getting stuck in the wrong local optima, especially since the policy structure is very sparse and asymmetric.

Upon getting a new demonstration $D$, the agent is updated with
\begin{align}
    \mathcal{D}_{i+1} &= \mathcal{D}_i \cup \{ D \} \\
    \mathcal{A}_{i+1} &= \big\langle \mathcal{I}_i, \mathcal{D}_{i+1}, \llbracket \alpha \rrbracket_i, \text{Train}(\llbracket \alpha \rrbracket_i, \mathcal{D}_{i+1}) \big\rangle
\end{align}
Similarly, for removing a demonstration, we have
\begin{equation}
    \mathcal{D}_{i+1} = \mathcal{D}_i \setminus \{ D \}
\end{equation}

\subsection{Policy Restructuring}

Given a set of instructions $\mathcal{I}_i$, we want to generate a policy structure $\llbracket \alpha \rrbracket_i$ that follows the instructions and performs well in the task domain.
As it is likely that a layperson would not fully specify all the details of the policy structure, we rely on the existing world knowledge in LLMs to make sensible design choices.
Essentially, the structure will be based primarily on the instructions $\mathcal{I}_i$, but the LLM is expected to fill in with its own knowledge and use its coding capability to translate it into an executable model.

To accommodate diverse instructions from laypeople, we increase robustness using chain-of-thought prompting \cite{wei2022chain} and in-context learning \cite{dong-etal-2024-survey}.
Specifically, the LLM is provided with an example of an instruction-model pair ($\mathcal{I}_\text{Lander}, \llbracket \alpha \rrbracket_\text{Lander}$) in a completely different task of Lunar Lander \cite{towers2024gymnasium}.
This makes use of the few-shot learning capability of LLMs and is expected to generate responses that follow the instruction better \cite{liu2024incomplete}.
The chain-of-thought prompting instructs the LLM to do the following steps:
\begin{enumerate}
    \item \textit{Extract all relevant features from the instructions.} This formalizes $V$ in the policy structure, resulting in a list of variable names and shapes such as
    {\small
    \begin{verbatim}
 * tiles (float32, shape=(B,L,7)) - input
 * tile_theta (float32, shape=(B,L))
 * raw_steer (float32, shape=(B,))
    \end{verbatim}
    \par
    }
    \vspace{-1em}
    \item \textit{Describe the structure in an English paragraph.} The generated paragraph adds filler content to the instructions.
    For example, the instruction might say only ``try to stay in the middle of the road", and the LLM needs to complete the components on how that translates into the action of steering. For instance, from the instruction above, the LLM generates:

    \begin{quote}
        \say{The policy computes summaries of the upcoming tiles (mean lateral offset and mean relative heading, ...). It builds a raw steering command as a linear combination of these summaries and the car's current heading. ...}
    \end{quote}
    \item \textit{Plan all the connections and operations.} This formalizes $P$ and $E$ in the policy structure.
    {\small
    \begin{verbatim}
* raw_steer (B,)
 - depends on mean_tile_x (positively correlated), 
     mean_tile_theta (negatively correlated), 
     current_heading (negatively correlated)
 - computed as w_x * mean_tile_x 
     + w_theta * mean_tile_theta 
     + w_head * current_heading
 - no bias (zero inputs -> zero steer)
 - w_x initialized positive (0.3), w_theta 
   initialized negative (-0.4), 
   w_head negative (-0.2)
    \end{verbatim}
    \par
  }
    \vspace{-1em}
    \item \textit{Generate the PyTorch model} by assembling all components.
    The generated model is executable and reflects the instructions. 
    An example is shown in Listing \ref{lst:pytorch}.
\end{enumerate}

\begin{lstlisting}[
  float,
  caption={Example of a PyTorch model generated},
  label={lst:pytorch},
  language=Python,
  style=mypython, 
  numbers=right,
  linewidth=\dimexpr\columnwidth-4mm\relax,
]
class RacecarPolicy(nn.Module):
  def __init__(self):
    super().__init__()
    self.w_x = nn.Parameter(
      torch.tensor(0.30, dtype=torch.float32)
    )
    self.w_theta = nn.Parameter(
      torch.tensor(-0.40, dtype=torch.float32)
    )
    # more initialization omitted
    
  def forward(self, tiles, indicators):
    # other details omitted
    raw_steer = (self.w_x * mean_tile_x) \
      + (self.w_theta * mean_tile_theta) \
      + (self.w_head * current_heading)
\end{lstlisting}

% \begin{lstlisting}[t]
% \caption{}
% \label{lst:pytorch}
% \footnotesize
% \begin{minted}[linenos, frame=single, escapeinside=||]{python}
% \end{minted}
% \end{lstlisting}

Having an intermediate step describing the structure in English has been empirically shown to improve the quality of the generation \cite{zhu2024bootstrapping}.
Therefore, the query process is
\begin{equation}
    \llbracket \alpha \rrbracket_i = \text{Restructure}(\mathcal{I}_i)
    \vcentcolon= LLM(\mathcal{S} + \mathcal{I}_\text{Lander} + \llbracket \alpha \rrbracket_\text{Lander} + \mathcal{I}_i)
\end{equation}
Where $+$ is message concatenation and $\mathcal{S}$ is the system prompt that explains the input and output format to LLM, chain-of-thought steps, task-agnostic specifications such as the differentiability constraints, etc.
% The Appendix provides all prompts used.

Upon receiving a new instruction $I$, the agent is updated with
\begin{align}
    \mathcal{I}_{i+1} &= \mathcal{I} \cup \{ I \} \\
    \llbracket \alpha \rrbracket_{i+1} &= \text{Restructure}(\mathcal{I}_{i+1}) \\
    \mathcal{A}_{i+1} &= \big\langle \mathcal{I}_{i+1}, \mathcal{D}_i, \llbracket \alpha \rrbracket_{i+1}, \text{Train}(\llbracket \alpha \rrbracket_{i+1}, \mathcal{D}_{i+1}) \big\rangle
\end{align}
Similarly, for removing an instruction, we have
\begin{equation}
    \mathcal{I}_{i+1} = \mathcal{I}_i \setminus \{ I \}
\end{equation}

\subsection{Strategy Summary and Rollouts}

During the structure generation process, an English description of the structure is already generated as an intermediate step, and it is closely tied to the generated PyTorch model.
Therefore, we can directly display it as a summary of the agent strategy.
The summary only reflects the structure of the policy along with some constants. 
It gives the user an idea of what features are being considered for which aspects of the task and how they are being connected.

For rolling out the policy, the user can pick whatever starting configuration they are interested in, and the agent deterministically selects an action based on its policy $\pi_{\langle \llbracket\alpha\rrbracket, \Theta\rangle}$ and keeps choosing actions based on the new observation received.
This process demonstrates the overall behavior of the agent.
% TODO: expand on this

\section{User Study}

To verify the efficacy of \interp{} among end-users without much technical background in policy training, we conducted an in-person between-subject user study with 34 users.
The study was approved by the Institutional Review Board of our institution.

\subsection{Setup}

\begin{figure}
    \centering
    \includegraphics[width=0.9\linewidth]{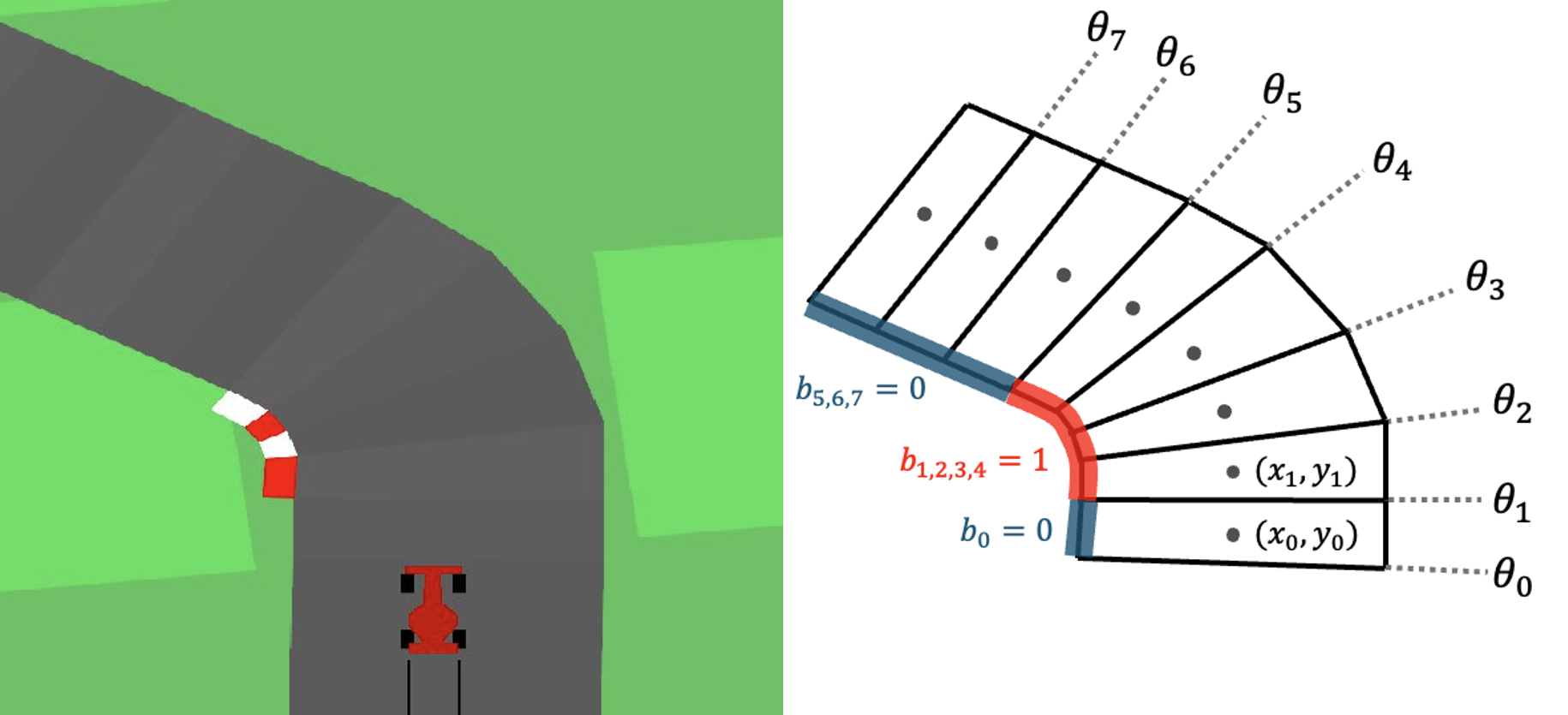}
    \caption{Environment rendering for the participant (left) and state representation for the policy (right). The coordinates for subsequent tiles are omitted due to space constraints.}
    \Description{
    The figure shows two images side by side.
    On the left it shows a red race car entering a corner on the race track curving left. The track consists of a sequence of tiles that each span from the left to the right of the track. The tile are color coded with light, medium, and dark gray. There is a red and white corner marker on the inside of the curve spanning four tiles long. The track is surrounded by green grass on the sides.
    On the right it shows an illustration of the contour of the eight tiles that constitutes the left turn.
    For each tile there is a line extending laterally with theta0 to theta7 annotated on the side.
    For each tile there is a gray dot annotating the center of the tile and for the first two tiles it is annotated with (x1, y1) and (x2, y2) respectively.
    On the left side there are red highlights on the inside of the curve where the corner marking is at on the image on the left.
    The red highlights are annotated with b123=1, and the tiles that are not part of the curve are highlighted with dark blue and b0456=0.
    }
    \label{fig:env}
\end{figure}

We use the car racing environment (Figure \ref{fig:env}) in the gymnasium package \cite{towers2024gymnasium} for our experiment. 
The task is chosen for two main reasons: 1) driving is a common task with which most people have prior knowledge, and 2) driving well on an unfamiliar simulator is hard for most laypeople, which reflects the challenge of giving optimal demonstrations in many real-world tasks.

We use a structured representation of the environment for the policy, where the observations consist of the current speed and heading of the egocentric race car and information about a fixed number of tiles in front of the race car (orientation, lateral distance, etc.).
The action space is the continuous control of the steering, gas pedal, and brake pedal.
The human participants see an image rendering of the environment and give demonstrations using a gamepad controller for continuous action inputs.

The main objective of the car racing task is to finish the track as quickly as possible.
The metric we use throughout the study is the effective average speed given a time cutoff $t_\text{cutoff}$
\begin{equation}
   EAS_{t_\text{cutoff}} = \dfrac{\min(N_\text{total}, N_\text{cutoff})}{\min(t_\text{total}, t_\text{cutoff})}
\end{equation}

Where $t_\text{cutoff}$ is the maximum time for the policy to run, $t_\text{total}$ is the total time used to complete the entire track, $N_\text{cutoff}$ is the number of tiles covered before the cutoff time, and $N_\text{total}$ is the total number of tiles on the track.
Consistent with the default setting \cite{towers2024gymnasium}, we use $t_\text{cutoff} = 1000$ steps ($40s$) and $N_\text{total} = 1000$.
We will use ``Average Speed" ($AS$) to represent $EAS_{1000}$ for the rest of the paper.

The study has two conditions: 1) the experimental condition, where participants are allowed to give instructions in addition to demonstrations and the model is trained with \interp{}, and 2) the baseline condition, where participants provide only demonstrations and the policy is represented by a multilayer perceptron (MLP) \cite{haykin1994neural}.
In both cases, the policy is trained with the same training configuration (learning rate, batch size, etc.)

\begin{figure}[t]
    \centering
    \includegraphics[width=\linewidth]{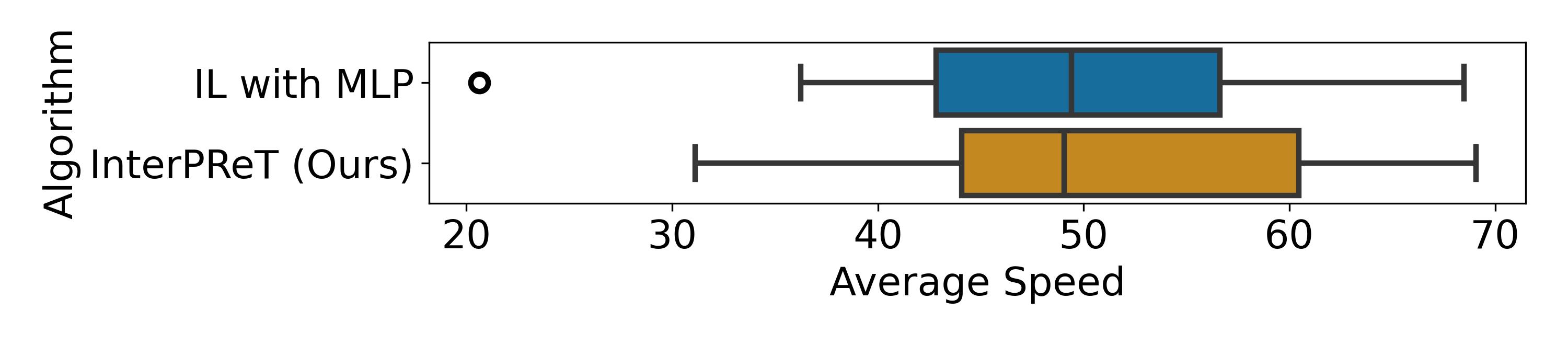}
    \caption{Average speed in nominal condition}
    \Description{
        A box plot where the x axis titles "average speed" ranging from 20 to 70, and the y axis titles algorithm with IL with MLP on the top row and InterPReT (Ours) on the bottom row.
        For the blue box on the top there is an outlier at around 21. The minimum whisker is at around 36, lower quartile is at 43, median at 49, upper quartile at 57, maximum at 68.
        For the orange box on the bottom the minimum whisker is at 31, lower quartile is at 44, median at 49, upper quartile at 60, maximum at 69.
    }
    \label{fig:res-speed-nominal}
\end{figure}

\subsection{Study Protocol}

$39$ participants were recruited through flyers and a posting in a designated research participants pool.
$4$ of the participants did not complete the study for personal or model-unrelated technical reasons, and $1$ participant was removed as their performance deviated more than three standard deviations from the general mean among all participants ($N=34, M=49.303, SD=13.581$).  
The removed participant had an average speed of $7.288$, and their trained policy was worse than a constant policy that only steered left (average speed $7.688$).  
% This left $34$ samples are balanced between the two conditions.
There are $19$ female participants and $15$ male participants, ranging from $18$ to $44$ years of age.
$29$ out of $34$ participants reported ``minimal or no experience in developing AI algorithms".
% More information on participants' background is in the Appendix.

Upon arrival, participants were instructed to fill out a consent form and a background survey before starting the study.
 Then a practice interface was shown to get them acquainted with controlling the race car with the gamepad controller.
In this stage, the track layouts are randomized so that the participant can practice general driving with the controller instead of overfitting to a specific track.
The participant had to cover $90\%$ of the track and achieve an average speed of $42$ to pass a lap. 
They had to pass $3$ laps consecutively to pass the practice session or attempt it $20$ times before they could move on.
This process took around $4$ minutes, on average, with a standard deviation of 3 minutes.

After practicing, the participants moved on to teaching the agent to drive.
The same track layout was used for both giving demonstrations and evaluating the learned policy's performance.
The participant could choose the initial position, orientation, and speed of the race car if they wanted to increase the diversity of their demonstrations or to see how well the learned policy performed under different conditions.
This allowed participants to give targeted demonstrations similar to DAgger \cite{ross2011reduction}.
During the study, we found that some participants took advantage of this setup and gave multiple individual demonstrations on different segments of the track because they were unable to reliably finish the entire track.

Participants were allowed to provide as many demonstrations (and instructions under the experimental condition) as they wanted, and there was no limit on the number of policies that they could train.
They could also test their trained agent with any starting configuration on the track for unlimited times.
However, they were required to test it at least four times to thoroughly evaluate their agent before stopping teaching.
To incentivize participants to try their best to train policies, we provided up to $\$5$ performance-based bonus, in addition to the $\$15$ base compensation, which depended on the average speed of their learned policy starting from rest and at an unknown position on the same track.
On average, participants received $\$18$ total compensation for both conditions.

After the participants submitted their final policy, they were instructed to complete a survey on system usability \cite{brooke1996sus}, the expectation of how the agent performs on different tracks, and some qualitative feedback on their experience.
During that time, their agent was automatically evaluated to determine their compensation.
At the end of the study, the participant was debriefed. 

\subsection{Hypotheses}

We hypothesize that having an instruction-instantiated structured policy helps to interpret demonstrations more effectively.
Therefore,
\begin{itemize}
    \itemindent=-15pt
    \item[] \textbf{H1}: \interp{} needs fewer demonstrations to achieve similar performance;
\end{itemize}
Furthermore, we hypothesize that given the structured policy space, \interp{} is more robust to different use cases beyond the ones that the participants trained in than the baseline.
Specifically, 
\begin{itemize}
    \itemindent=-15pt
    \item[] \textbf{H2}: \interp{} performs better on \textit{unseen} tracks;
    \item[] \textbf{H3}: \interp{} performs better in \textit{edge cases} starting configurations on the \textit{seen} training track;
    \item[] \textbf{H4}: \interp{} performs better on the \textit{seen} training track with \textit{action noise};
\end{itemize}
Also, we collect user's perceived performance to complement the single-factored quantitative evaluation. 
As the instructions complement potentially imperfect demonstrations, we hypothesize that
\begin{itemize}
    \itemindent=-15pt
    \item[] \textbf{H5}: Users perceive \interp{} to perform better than them;
\end{itemize}
Finally, since most laypeople can think of some general instruction for driving easily, we hypothesize that 
\begin{itemize}
    \itemindent=-15pt
    \item[] \textbf{H6}: The usability of \interp{} is \textit{no worse} than the baseline.
\end{itemize}

\section{Quantitative Results}

\subsection{Data Usage for Training}

\begin{figure}[t]
    \centering
    \includegraphics[width=\linewidth]{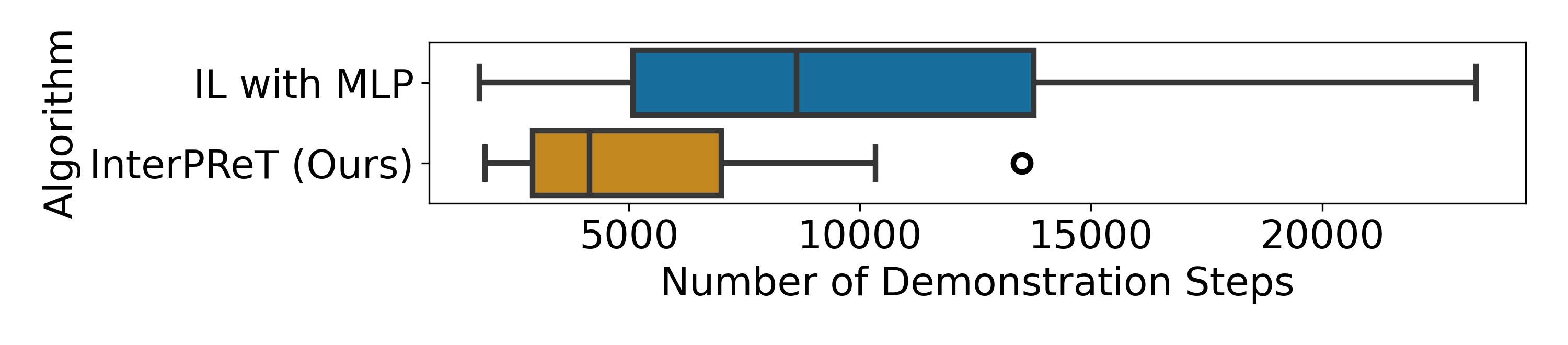}
    \caption{Number of demonstrations used}
    \Description{
        A box plot where the x axis titles "number of demonstration steps" ranging from 1500 to 24000, and the y axis titles algorithm with IL with MLP on the top row and InterPReT (Ours) on the bottom row.
        For the blue box on the top the minimum whisker is at around 1759, lower quartile is at 5093, median at 8628, upper quartile at 13763, maximum at 23326.
        For the orange box on the bottom the minimum whisker is at 1882, lower quartile is at 2916, median at 4149, upper quartile at 6996, maximum at 10500, and an outlier at 13497.
    }
    \label{fig:res-demonstrations}
\end{figure}

Figure \ref{fig:res-speed-nominal} shows the average speed of the trained policies in nominal configuration (seen track, no noise, starting at the center of the track with initial speed $40$).
The Welch t-test \cite{welch1938significance} shows no evidence suggesting that the mean of \interp{} ($n = 17, M = 51.488, SD = 11.119$) differs from the baseline ($n = 17, M = 49.591, SD = 12.364$) with $t(31.65) = -0.470, p = 0.641$.
A further two one-sided test procedure \cite{schuirmann1987comparison} indicates that the two groups are equivalent with bounds $\pm 9$ ($p=0.0439$).
This verifies that participants from both conditions have similar stopping criteria and achieve similar performance in nominal configurations.

Figure \ref{fig:res-demonstrations} shows the number of demonstrations given by the two groups to achieve that performance.
The Welch t-test shows that \interp{} requires fewer demonstrations ($n = 17, M = 5454.529, SD = 3315.882$) than baseline ($n = 17, M = 9699.118, SD = 6233.427$) with significance ($t(24.38) = 2.479, p = 0.020$).
This supports \textbf{H1} that \interp{} is more sample efficient.

\subsection{Robustness of Trained Policies}

\begin{figure}[t]
    \centering
    \includegraphics[width=\linewidth]{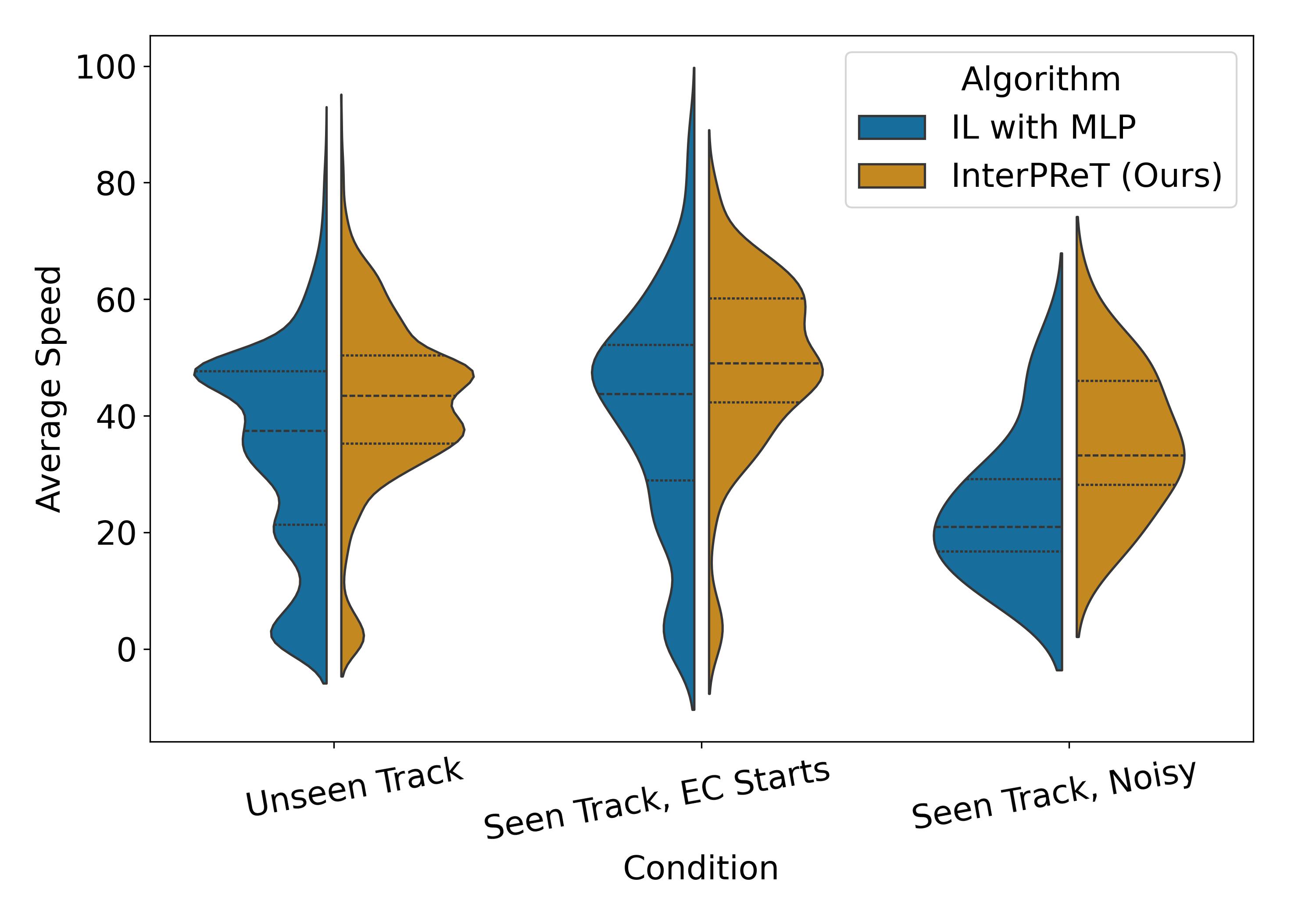}
    \caption{Average speed per run across different conditions}
    \Description{
        Three split violin plots showing the distribution of the two algorithms in three conditions.
        For each violin plot there is blue part on the left denoting IL with MLP and an orange part on the right denoting InterPReT (Ours).
        The x axis is three conditions: unseen track, seen track with edge case starts, and seen track with noise.
        The y axis is average speed.
        For the unseen condition on the left, the baseline violin plot shows four peaks, each at 50, 38, 20, and 5 respectively with the one at 50 being the most intense, 38 being the second, 0 being the third and 20 being the fourth. The inside of the violin plot annotates the quantiles where the lower quartile is around 20, the median is at 39 and the upper quartile is at 50 coinciding the largest peak.
        The orange violin plot shows three peaks each at 45, 37, and 5 respectively with the first two being considerably more intense than the last.
        The lower quartile is around 38, the median is around 42, and the upper quartile is around 50.
        Both violin plots span from -5 to 98.
        For the seen track with edge cases start condition in the middle, the baseline violin plot shows two peaks, each at 48 and 0 respectively with the one at 48 being much more intense. It shapes like a normal curve. The inside of the violin plot annotates the quantiles where the lower quartile is around 30, the median is at 42 and the upper quartile is at 56.
        It spans from -8 to 100.
        The orange violin plot shows three peaks each at 60, 45, and 0 respectively with the first two being considerably more intense than the last.
        The lower quartile is around 42, the median is around 48, and the upper quartile is around 60.
        It spans from -5 to 90.
        For the seen track with noise condition on the right, the baseline violin plot shows a left skewed normal where the peach is at 19. 
        It spans -3 to 70 with lower quartile at 18, median at 20, and upper quartile at 36.
        The orange violin plots shows a normal curve where the peak is at 35.
        It spans 3 to 75 with lower quartile at 36, median at 38, and upper quartile at 45.
    }
    \label{fig:res-speed-per-run}
\end{figure}

To test for the final learned policies' robustness, we took the submitted policy for each participant and evaluated them in different conditions.
Specifically, the ``Unseen Track" condition consisted of $10$ unseen tracks, each with $47$ starting configurations that varied the initial speed, orientation, and position of the race car;
the ``Seen Track, Edge Case Starts" condition consisted of $46$ edge-case starting configurations on the training track;
and the ``Seen Track, Noisy" condition started with a nominal initial position but introduced action noise (at $2$ levels) to the policy rollout.
This resulted in a table of shape $N \times (10 \cdot 47 + 46 + 2)$.

We used a linear mixed-effect model \cite{lindstrom1988newton} where algorithm and condition are the main effects, and participant ID and track ID are the random effects:
\begin{equation}
   AS \sim \beta_0 + \beta_1\cdot Algo + \beta_2\cdot Cond + \beta_3\cdot(Algo \times Cond) + u_p + v_t + \epsilon
\end{equation}

\noindent where $Algo$ is categorical with $2$ values ($0$ for the baseline and $1$ for \interp{}), $Cond$ is categorical with $3$ values (three evaluation conditions), $\beta_*$ are linear coefficients, $u_p$ is the intercept for each participant, $v_t$ is the intercept for each track, and $\epsilon$ is noise.
This accounts for the differences in the participants' proficiency in using the controller and the difficulty of each track.
We use $\beta_{31}$ to represent the coefficient of \interp{} in ``Seen Track, Edge Case Starts" condition and $\beta_{32}$ for \interp{} in ``Seen Track, Noisy" condition.

Using the Wald test for contrasts \cite{wald1943tests} and Holm-Sidek correction \cite{vsidak1967rectangular}, the results (Figure \ref{fig:res-speed-per-run}) \textbf{support all hypotheses H2-4}: 
\interp{} outperforms the baseline in ``Different Tracks" (contrast $\hat\beta=8.096, SE=3.196, CI_{95\%}=[1.833, 14.359], z=2.533, p=0.0113, p_\text{Holm-Sildek} = 0.0335$), 
in ``Same Track, Edge Case Starts" (contrast $\hat\beta_1 + \hat\beta_{31} = 8.461, SE=3.839, CI_{95\%}=[0.936, 15.985], z=2.204, p=0.0275, p_\text{Holm-Sildek}=0.0401$), 
and in ``Same Track, Noisy" (contrast $\hat\beta_1 + \hat\beta_{32} = 11.027, SE = 4.788, CI_{95\%}=[1.644, 20.411], z=2.303, p=0.0212, p_\text{Holm-Sildek}=0.0421$).
This shows that \interp{} is significantly better than the baseline in all conditions.

\subsection{User Perception}

\begin{figure}[t]
    \centering
    \includegraphics[width=\linewidth]{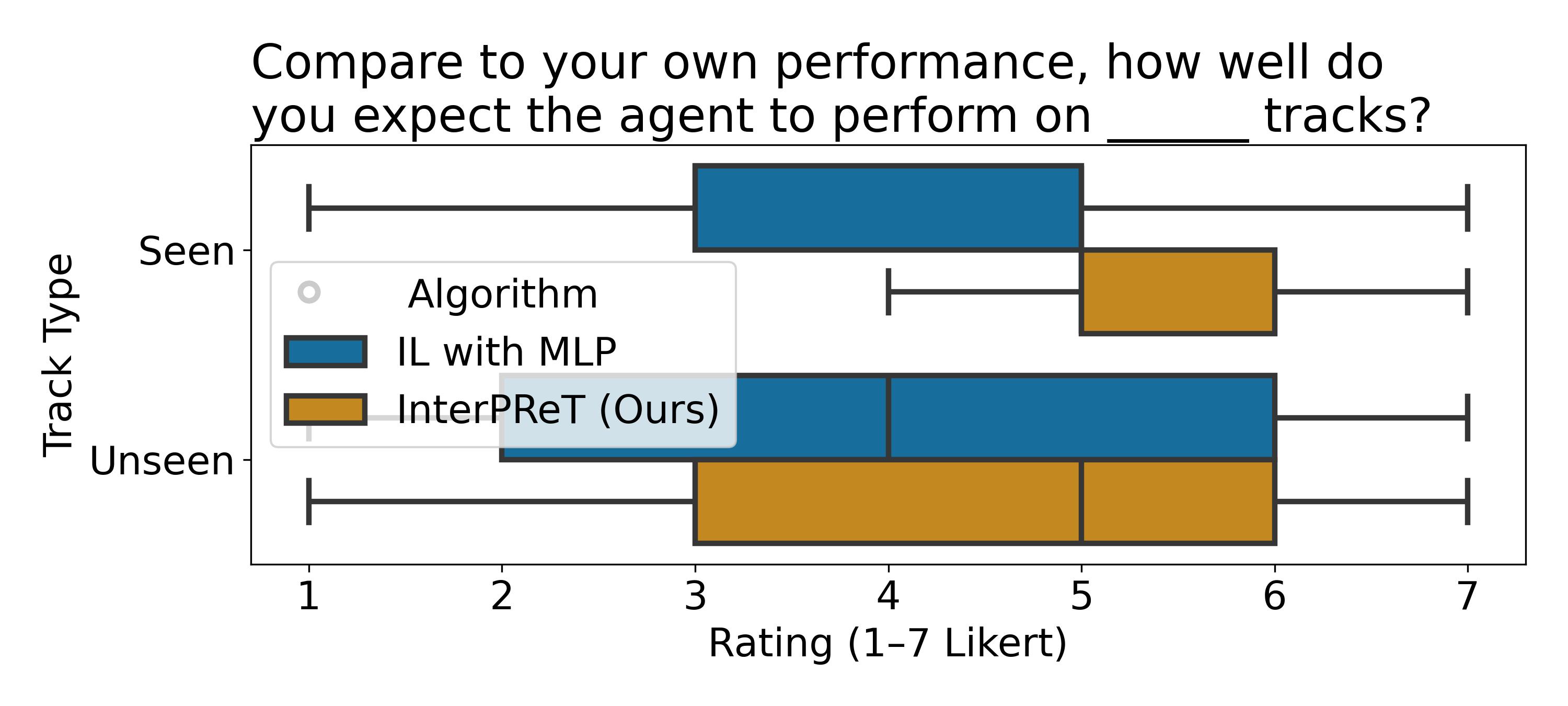}
    \caption{Users' expectation for the trained AI}
    \Description{
        The figure is titled compare to your own performance, how well do you expect the agent to perform on blank tracks.
        It shows boxplot with four rows, corresponding to baseline on seen tracks, interpret on seen tracks, baseline on unseen tracks, and interpret on unseen tracks from top to bottoms.
        The x axis titles "Rating (1-7 Likert)".
        The first box shows a minimum of 1, lower quartile at 3, median at 5, upper quartile at 5, and maximum at 7.
        The second box shows an outlier at 1, minimum at 4, lower quartile at 5, median at 5, upper quartile at 6, and maximum at 7.
        The third box shows a minimum of 1, lower quartile at 2, median at 4, upper quartile at 6, and maximum at 7.
        The fourth box shows a minimum at 1, lower quartile at 3, median at 5, upper quartile at 6, and maximum at 7.
    }
    \label{fig:res-expectation}
\end{figure}

Figure \ref{fig:res-expectation} shows the' perceived performance of the trained policy by the users.
For the seen track, the \interp{} result is mostly positive with an outlier who rated $1$, so the distribution is highly skewed and does not meet the normality assumption (Shapiro-Wilk normality test returns $p=0.0019$)
Therefore, a Mann-Whitney U test is used to compare the two distributions and shows that user perception in the \interp{} condition ($n = 17, M = 5.353, Mdn = 5.000, SD = 1.412$) is higher than baseline ($n = 17, M = 4.294, Mdn = 5.000, SD = 1.611$) with significance ($U = 86.0, p = 0.0393$).
However, for unseen tracks, users' perception of the learned policy is similar between \interp{} ($n = 17, M = 4.529, SD = 2.004$) and baseline ($n = 17, M = 3.941, SD = 1.952$), and there is no evidence suggesting that there is a difference ($t(14.16) = -0.979, p = 0.344$) based on a Welch t-test.
This shows \textbf{partial support for H5}.  In particular, it shows that although users recognize the better performance of policies trained by \interp{}, they cannot identify the key features that allow it to be generalized to other unseen scenarios.

\begin{figure}[t]
    \centering
    \includegraphics[width=\linewidth]{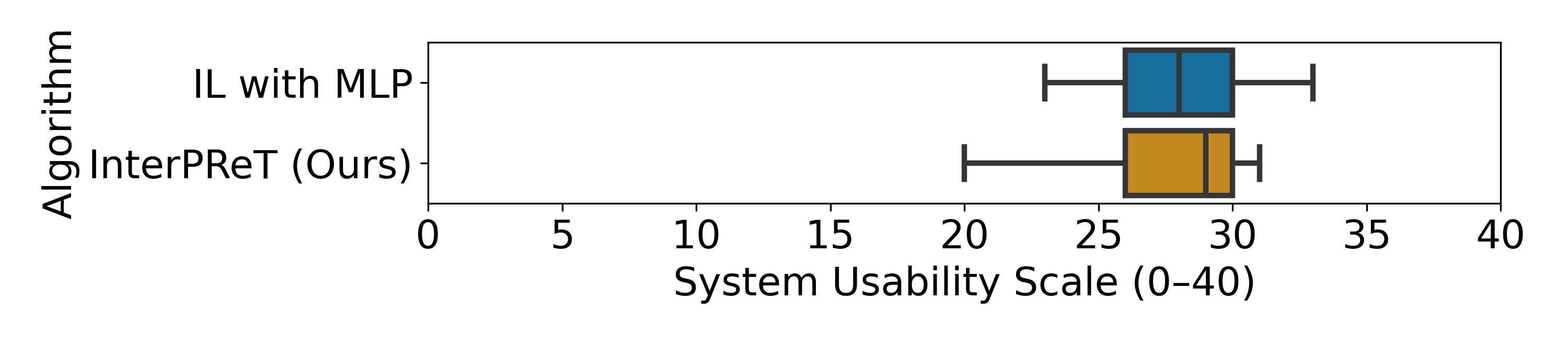}
    \caption{Users' rating on system usability}
    \Description{
        A box plot where the x axis titles "system usability scale (0-40)" ranging from 0 to 40, and the y axis titles algorithm with IL with MLP on the top row and InterPReT (Ours) on the bottom row.
        For the blue box on the top the minimum whisker is at around 23, lower quartile is at 26, median at 28, upper quartile at 30, maximum at 33.
        For the orange box on the bottom the minimum whisker is at 20, lower quartile is at 26, median at 29, upper quartile at 30, maximum at 31.
    }
    \label{fig:res-sus}
\end{figure}

A one-sided Welch t-test \textbf{confirms H6}: \interp{} ($n = 17, M = 27.765, SD = 2.969$) is not worse than the baseline ($n = 17, M = 27.824, SD = 2.856$) with significance ($t(31.95) = 0.059, p_\text{one-sided} = 0.0235$) on the system usability scale (SUS) \cite{brooke1996sus} reported by the participants (Figure \ref{fig:res-sus}).
Additionally, no evidence suggests a difference in completion times (in minutes) between \interp{} ($M = 35.919, SD = 12.082$) and baseline ($M = 28.489, SD = 14.138$ minutes), $t(31.24) = -1.647, p = 0.110$.
This shows that despite having to think about and articulate some driving instructions, it did not affect the overall usability of the teaching system.

\section{Analysis}

\subsection{Improvement through Interactions}

\begin{figure}[t]
    \centering
    \includegraphics[width=0.9\linewidth]{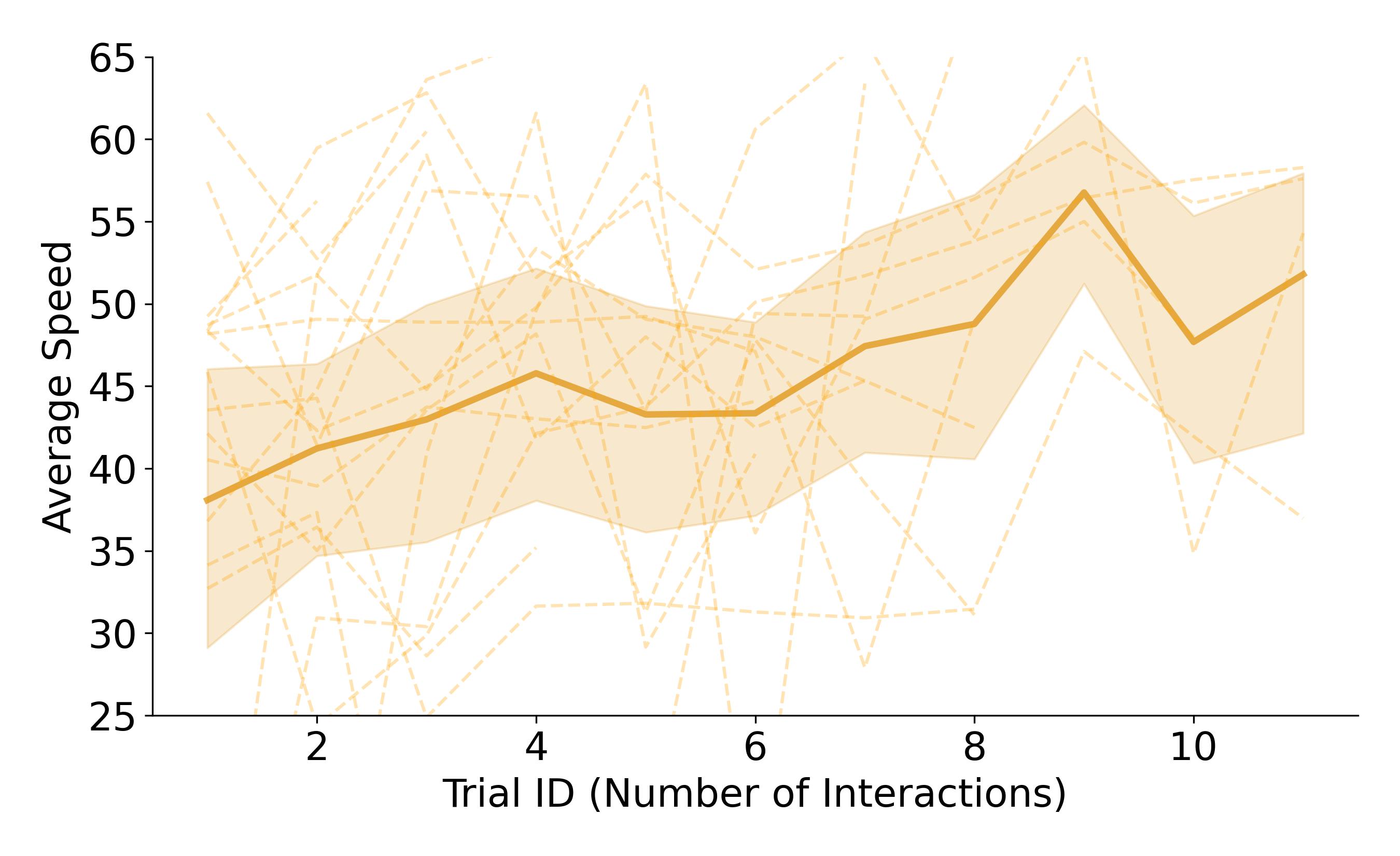}
    \caption{
        Performance increase among participants using \interp{}. 
        Shaded area denotes $95\%$ confidence interval.
    }
    \Description{
        A spaghetti plot showing the trend of the policy trained by each participants over time overlayed with line plot of the the mean across participants.
        The x axis titles "trial ID" ranging from 1 to 11 and the y axis titles average speed ranging from 25 to 65.
        The line goes from roughly 38 at trial ID 1 to 51 at trial ID 11 with a peak at average speed of 56 at trial ID 8 and a dip at average speed 45 at trial ID 10.
        The shaded area is approximated 7.5 width throughout the line plot.
    }
    \label{fig:res-trend}
\end{figure}

We evaluated \textit{all} the policies trained by each participant using \interp{} on the seen track at the nominal starting position (Figure \ref{fig:res-trend}).
To test whether the improvement has statistical significance, we fit another linear mixed-effect model with the index of each trained policy as $TrialID$:
\begin{equation}
   AS \sim \beta_0 + \beta_1\cdot TrialID + u_p + \epsilon
\end{equation}
The Wald test shows that \interp{}' performance improves as the number of interactions increases (contrast $\hat\beta_1=1.212, SE=0.412,$ $CI_\text{95\%}=[0.404, 2.020], z=2.940, p=0.00328$).
This shows that interaction is important for policy training.
Intuitively, this attributes to the users being able to give targeted demonstrations or instructions after receiving feedback from the agent.

\subsection{Handling Diverse Instructions}

\begin{figure}[t]
    \centering
    \includegraphics[width=\linewidth]{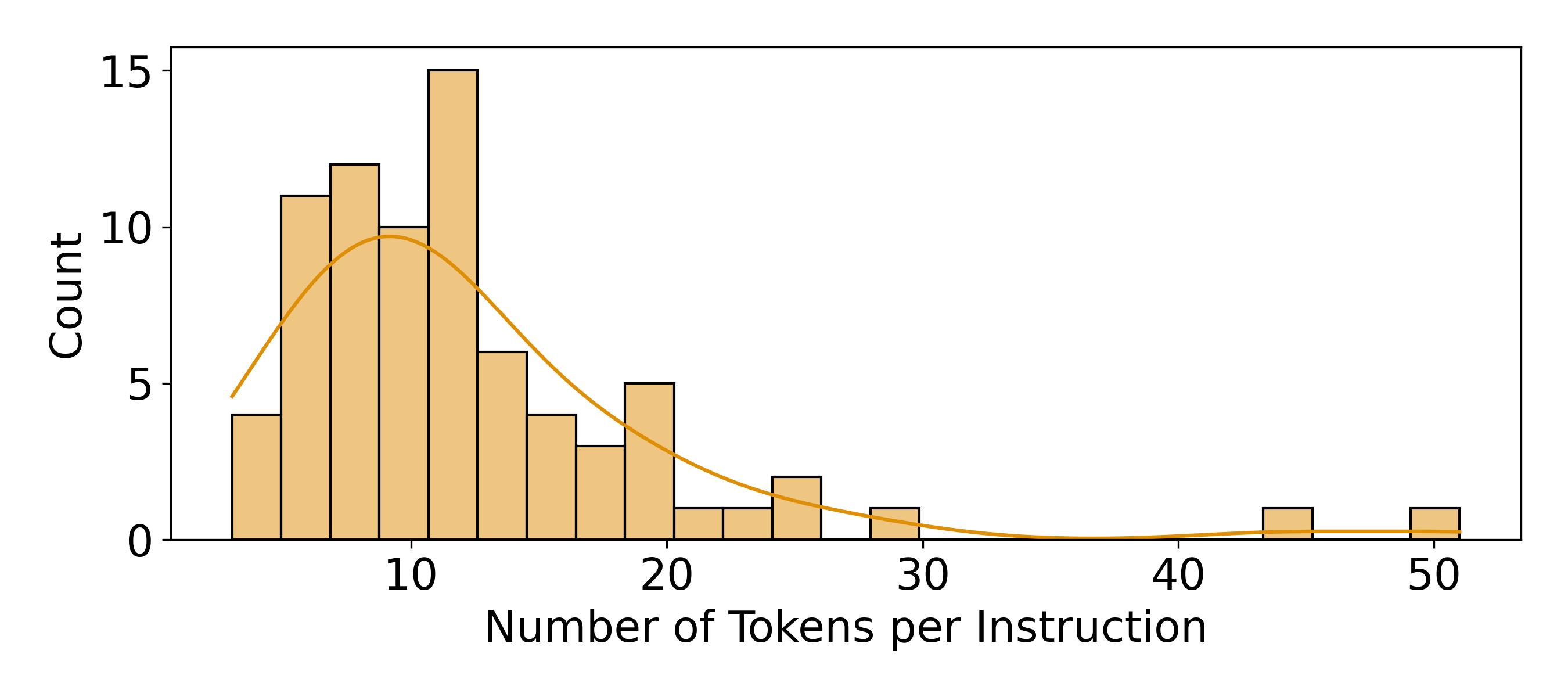}
    \caption{ Distribution of instruction lengths }
    \Description{
    A histogram showing the distribution of number of tokens per instruction.
    The x axis titles number of tokens per instruction and y axis titles counts.
    The distribution is left skewed with the tallest group at 10-12 at count of 15.
    The KDE curve peaks at 10.
    All instructions have token length less than 30 except for one at 44 and another at 51.
    }
    \label{fig:token-count}
\end{figure}

\begin{lstlisting}[
  style=mypython, 
  float, 
  caption={Examples of latent variables generated}, 
  label={lst:stay-on-road},
  linewidth=\dimexpr\columnwidth-1mm\relax,
]
# (irrelevant instructions and code omitted)
# Participant A
# Instruction: never leave the road
denom = |\higreen{self.track\_half\_width}| + 1e-6
off_track = |\hired{torch.clamp}|(
  (|\hiblue{abs\_x}| - |\higreen{self.track\_half\_width}|) / denom, 0.0, 1.0
)

# Participant B
# Instruction: move car forward, keeping car on black race
# track
offtrack = |\hired{torch.clamp}|(
  (|\hiblue{torch.abs(closest\_x)}| - |\higreen{self.track\_half\_width}|) 
  / |\higreen{self.track\_half\_width}|, min=0.0, max=1.0
)

# Participant C
# Instruction: If on the edge of the track, move over so as
#   to not go off the track
edge_proximity = |\hired{torch.relu}|(
  |\hiblue{torch.abs(nearest\_x)}| - |\higreen{self.track\_half\_width}|
)
\end{lstlisting}

% \begin{listing}[ht]
% \caption{Examples of latent variables generated}
% \label{lst:stay-on-road}
% \footnotesize
% \begin{minted}[linenos, frame=single, escapeinside=||]{python}
% # (irrelevant instructions and code omitted)
% ### Participant A ###
% # Instruction: never leave the road
% denom = |\higreen{self.track\_half\_width}| + 1e-6
% off_track = |\hired{torch.clamp}|(
%   (|\hiblue{abs\_x}| - |\higreen{self.track\_half\_width}|) / denom, 0.0, 1.0
% )
% 
% ### Participant B ###
% # Instruction: move car forward, keeping car on black race track
% offtrack = |\hired{torch.clamp}|(
%   (|\hiblue{torch.abs(closest\_x)}| - |\higreen{self.track\_half\_width}|) 
%   / |\higreen{self.track\_half\_width}|, min=0.0, max=1.0
% )
% 
% ### Participant C ###
% # Instruction: If on the edge of the track, move over so as to not 
% #   go off the track
% edge_proximity = |\hired{torch.relu}|(
%   |\hiblue{torch.abs(nearest\_x)}| - |\higreen{self.track\_half\_width}|
% )
% \end{minted}
% \end{listing}

\begin{lstlisting}[
  style=mypython, 
  float, 
  caption={Examples of control methods generated}, 
  label={lst:slow-on-turns},
  linewidth=\dimexpr\columnwidth-1mm\relax,
]
# Participant D
# Instruction: Brake proportional to speed when a curve with
#   red and white barrier is ahead
barrier_strength = border_ahead.mean(dim=1)
brake_raw = self.w_brake * speed * barrier_strength

# Participant E
# Instruction: At the corners try to slow down and decelerate 
#   then drive at the previous speed
target_speed = torch.where(
  corner_any > 0.0,
  self.v_target_corner,  self.v_target_straight
)
speed_error = target_speed - v
brake_raw = torch.relu(self.k_brake * (-speed_error))
\end{lstlisting}

% \begin{listing}[ht]
% \caption{Examples of control methods generated}
% \label{lst:slow-on-turns}
% \footnotesize
% \begin{minted}[linenos, frame=single, escapeinside=||]{python}
% ### Participant D ###
% # Instruction: Brake proportional to speed when a curve with red 
% #   and white barrier is ahead
% barrier_strength = border_ahead.mean(dim=1)
% brake_raw = self.w_brake * speed * barrier_strength
% 
% ### Participant E ###
% # Instruction: At the corners try to slow down and deccelerate 
% #   then drive at the previous speed
% target_speed = torch.where(
%   corner_any > 0.0,
%   self.v_target_corner.expand(B), self.v_target_straight.expand(B)
% )
% speed_error = target_speed - v
% brake_raw = torch.relu(self.k_brake * (-speed_error))
% \end{minted}
% \end{listing}

LLMs have been shown to sometimes be very sensitive to prompt changes - minor paraphrasing of prompts may lead to drastically different outputs \cite{salinas2024butterfly}.
Therefore, it is important to investigate how robust it is in handling the diverse languages used by participants.

Figure \ref{fig:token-count} shows the distribution of the number of tokens in the instructions given by the participants.
On average, each user uses $55$ tokens, which is a $90\%$ reduction from previous work that requires detailed instructions \cite{zhu2025sample}.
This highlights the user friendliness of \interp{}.
The provided instructions cover a wide range of styles:
\begin{itemize}
    \itemindent=-15pt
    \item[] \textit{\textbf{Terse and Abstract}}: \say{Stay within the grey track}
    \item[] \textit{\textbf{Terse and Specific}}: \say{Desired speed is 70.}
    \item[] \textit{\textbf{Verbose}}: \say{when turning prioritize the middle of the road rather than the inside of the bend. this will limit your chances of hitting grass. in general, try to stay in the middle of the road since you are surrounded by grass}
    \item [] \textit{\textbf{Second-person}}: \say{speed up to go as fast as you can}
    \item [] \textit{\textbf{Third-person}}: \say{Keep the car straight on a straight road}
    \item [] \textit{\textbf{Typo}}: \say{Turn a corner}
\end{itemize}
Except for a few common instructions such as ``stay on the road" and ``slow down at turns", most instructions appear only once.

Listing \ref{lst:stay-on-road} shows examples of the generated models having the same latent variable representation from syntactically different instructions.
This shows that \interp{} can correctly identify key latent features and is robust to variations in the expression of instructions.
Meanwhile, Listing \ref{lst:slow-on-turns} shows that the generated model correctly mirrors the user's instruction of ``brake proportional to ..." by using a multiplication function directly on the output action space (participant D), and it also explicitly modeled the desired speed when the instruction focuses on the speed of the car instead of a specific action (participant E).
This shows that it can capture the nuances and follow the instructions closely instead of returning the same model every time.

\subsection{User Leveraging Strategy Summary}

We found that some participants deliberately referenced the variable names reflected in the summary of the strategy to communicate better with the model generation process.
For example, after the agent displayed:

\begin{quote}
    \say{... It [the agent] computes a desired speed that is low (25) when a sharp corner (\hired{border=1}) is close, and high (85) when no corner is near. ...}
\end{quote}
The user used the same concept of \hired{border} to define straight lines in their instructions:

\begin{quote}
    \say{Accelerate to maximum speed if the road is straight or the corner is not sharp (\hired{border<0.4}) ahead}
\end{quote}
Which is generated as

\tagpdfsetup{para/tagging=false}
\begin{lstlisting}[style=mypython, frame=none, numbers=none]
target_speed = torch.where(
  |\hired{border\_max < 0.4}|, 
  torch.ones_like(base_target), base_target
)
\end{lstlisting} \par
\tagpdfsetup{para/tagging=true}

This shows that in addition to providing inspiration for new demonstrations and instructions, the summary generated establishes a common language that is shared between the user and the agent such that the user can have direct control over the policy structure even if they have no coding experience.

\subsection{Additional Ablations}

\begin{figure}
    \centering
    \includegraphics[width=1.0\linewidth]{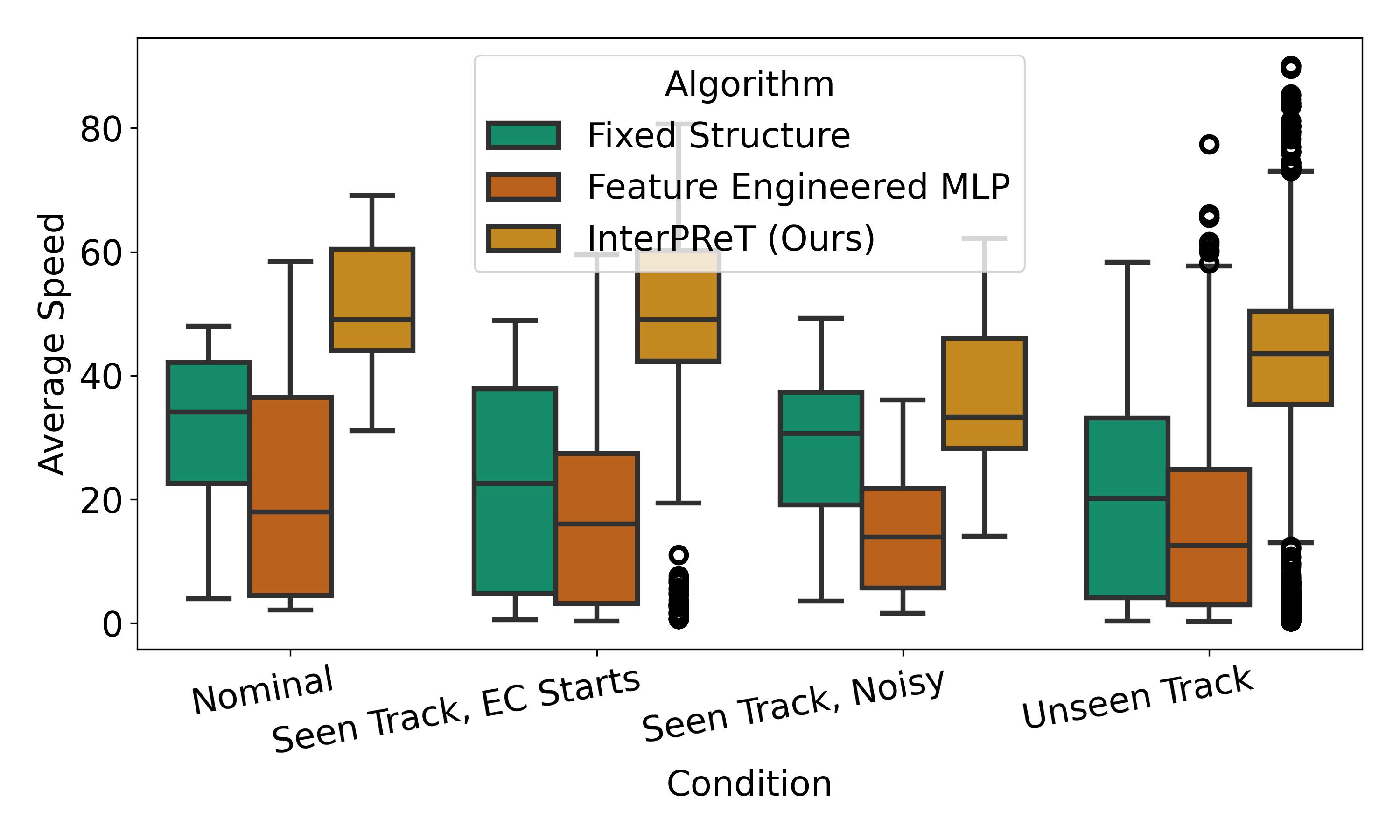}
    \caption{Average speed for different algorithm variations}
    \Description{
        Box plots showing four conditions and three algorithms.
    }
    \label{fig:res-ablation}
\end{figure}

In this work, language feedback and structure generation are treated as a whole as the structures are based on the language instructions.
To illustrate this, we use the data collected from the \interp{} group and compare our method with two alternative formulations:
\begin{itemize}
    \item Fixed Structure (no instruction): A policy structure generated by LLM without any human instructions.
    \item Feature-Engineered MLP (no structure): The LLM generates only a feature-engineering layer, based on user instructions, before the fully connected layers in an MLP.
\end{itemize}

Figure \ref{fig:res-ablation} shows the average speed per participant in each condition for \interp{} and the two alternatives.
Paired Welch t-test with Holm-Sidak correction shows that \interp{} outperforms the others in all conditions ($p_\text{Holm-Sildek} = 0.0259$ for Fixed Structure, Seen Track, Noisy; $p_\text{Holm-Sildek} < 0.001$ for the rest).

This shows that having a structured policy alone, without user instructions, is unsuitable for imitation learning, as the LLM may have modeling assumptions that misalign with the demos.
Additionally, users may provide an incomplete set of features, and the unstructured policy cannot efficiently interpret the demos.

\section{Discussion}

\subsection{Pathway to Deployment}

% The study setup reflects that after the user gets a new household robot, they can place it in different places (the user picks the starting pose freely) in their house to teach and test the robot's customized skills, but they cannot exhaustively train the robot in every scenario (e.g., getting new furniture, having unseen guests, etc.).
% The test settings imitate when the user rearranges their house (unseen tracks), re-positions the robot (different starting configurations), and if the robot's actuators degrade over time (action noise).
% As the setup is closely related to the real-world environments, we expect it to showcase the same robustness when deployed to a physical robot.
% Additionally, since the agent can learn low-level actions, it could be applied to more complex domains such as mobile manipulation for household cooking and cleaning tasks.

To deploy the current approach to the real world, we would need to convert the raw perception into a structured representation by using models such as lane marker detection \cite{huang2023anchor3dlane}, object segmentation \cite{carion2025sam3segmentconcepts}, etc.
This allows LLMs to make use of the semantic information of the features in the observations.
To collect the expert demonstrations, we would need teleoperation devices that fit the robots' morphology (e.g., \cite{fu2024mobile, liu2025factr}).
Once those systems are integrated, we can deploy our approach to a physical robot in the real world.
Since the agent can learn low-level actions, we anticipate that it can perform physical tasks like opening doors or sweeping, where there is strong semantics and easy-to-describe motions.
Usability can also be increased by integrating bidirectional language feedback \cite{wang2025cori} and using a local LLM \cite{agarwal2025gpt} for lower latency.

\subsection{Limitation and Future Work}

% The current approach generates the model structure from instructions in an open-loop fashion and is unaware of the potential mismatch between the generated model and the user demonstrations.
% For example, when we included the angular velocity in the observation space during our pilot testing, GPT generated a learnable parameter between the steering output and the current angular velocity.
% From the model design standpoint, this weight was supposed to be negative, acting as a damping effect of the race car's wobbliness.
% However, in the demonstrations where the race car passes through turns, the steering direction is the same as the angular velocity, leading the model to learn a positive value for the weight, making the race car more wobbly.
% To solve this, one future direction is to close the loop and feed the trained agent's behavior back to GPT to inform better structure generation, similar to \cite{ma2023eureka}.

\interp{} learns all policy parameters entirely from the demonstrations and treats all the demonstrations equally.
The instructions are only used to help interpret the user demonstrations as opposed to augmenting their demonstrations.
For example, if the user says ``go fast" while giving a relatively slow demonstration, then the agent is going to assume that their speed is ``fast".
The agent is not capable of incorporating instructions like ``go faster than my demonstrations".
Future work could explore instruction-guided reinforcement learning \cite{golchha2024language} to improve agents' performance beyond user demonstrations.

Although \interp{} provides more interpretability of its model structure than generic unstructured models, its feedback to the user is still very primitive.
Future work could investigate more expressive explanations (e.g., by integrating visuals and animations), giving verbal explanations that are grounded in the rollouts, suggesting more informative starting configurations \cite{lee2025improving}, or using second-order theory of mind \cite{callaghan2025second} to give user-targeted clarifications.

% Additionally, the current approach does not ground the demonstrations to the instructions.
% It would provide the user with more flexibility if they can specify certain demonstrations are specifically to show the point of certain instructions.

\section{Conclusion}

This paper introduced \interp{}, an interactive teaching paradigm where the agent restructures its policy structure according to  user instructions and trains the policy weights based on user demonstrations.
Through a user study with laypeople, we show that our proposed method achieves better robustness in unseen environment settings, receives better perceived performance, and does not affect the overall usability of the system.
We also analyze the generated models and confirmed that the components, such as multiround interactions and summary generation, all contribute to the learning agents' performance gain, and the system is able to handle a diverse variety of user instructions.
And finally, we explain the limitations of the current approach and lay out the roadmap for future work and eventual deployment in society.

% \interp{} allows laypeople to teach new customized skills to the agent with good generalizability.
% This means that anyone, even without technical literacy, can teach the agent how to do things specifically in their home, and does not have to worry about the agent making a mess in unseen scenarios.
% In doing so, \interp{} lowers the barrier to personalized robots and paves the way for a future where robots empower individuals and enrich communities.

\newpage
\begin{acks}
We would like to thank Shridhula Srinivasan and Justin Ma for their help in prototyping the user interface, experimenting with prompt engineering, and running some pilot user studies. 
Additional thanks to Justin for running some final studies. 

This research has been partially supported by Microsoft Corporation as part of the Keio CMU partnership. And Feiyu is supported by the Softbank Group - Arm PhD Fellowship.
\end{acks}


\balance
\bibliographystyle{ACM-Reference-Format}
\bibliography{sample-base}

@article{agarwal2025gpt,
  title={gpt-oss-120b \& gpt-oss-20b model card},
  author={Agarwal, Sandhini and Ahmad, Lama and Ai, Jason and Altman, Sam and Applebaum, Andy and Arbus, Edwin and Arora, Rahul K and Bai, Yu and Baker, Bowen and Bao, Haiming and others},
  journal={arXiv preprint arXiv:2508.10925},
  year={2025}
}

@inproceedings{ayalew2025enabling,
  title={Enabling End Users to Program Robots Using Reinforcement Learning},
  author={Ayalew, Tewodros W and Wang, Jennifer and Littman, Michael L and Ur, Blase and Sebo, Sarah},
  booktitle={2025 20th ACM/IEEE International Conference on Human-Robot Interaction (HRI)},
  pages={767--777},
  year={2025},
  organization={IEEE}
}

@inproceedings{babe2024studenteval,
  title={Studenteval: A benchmark of student-written prompts for large language models of code},
  author={Babe, Hannah McLean and Nguyen, Sydney and Zi, Yangtian and Guha, Arjun and Feldman, Molly Q and Anderson, Carolyn Jane},
  booktitle={Findings of the Association for Computational Linguistics: ACL 2024},
  pages={8452--8474},
  year={2024}
}

@inproceedings{brawer2023interactive,
  title={Interactive policy shaping for human-robot collaboration with transparent matrix overlays},
  author={Brawer, Jake and Ghose, Debasmita and Candon, Kate and Qin, Meiying and Roncone, Alessandro and V{\'a}zquez, Marynel and Scassellati, Brian},
  booktitle={Proceedings of the 2023 ACM/IEEE International Conference on Human-Robot Interaction},
  pages={525--533},
  year={2023}
}

@article{brooke1996sus,
  title={SUS-A quick and dirty usability scale},
  author={Brooke, John and others},
  journal={Usability evaluation in industry},
  volume={189},
  number={194},
  pages={4--7},
  year={1996},
  publisher={London, England}
}

@article{cao2021learning,
  title={Learning from imperfect demonstrations from agents with varying dynamics},
  author={Cao, Zhangjie and Sadigh, Dorsa},
  journal={IEEE Robotics and Automation Letters},
  volume={6},
  number={3},
  pages={5231--5238},
  year={2021},
  publisher={IEEE}
}

@article{callaghan2025second,
  title={Second-order Theory of Mind for Human Teachers and Robot Learners},
  author={Callaghan, Patrick and Simmons, Reid and Admoni, Henny},
  journal={arXiv preprint arXiv:2503.16524},
  year={2025}
}

@misc{carion2025sam3segmentconcepts,
      title={SAM 3: Segment Anything with Concepts},
      author={Nicolas Carion and Laura Gustafson and Yuan-Ting Hu and Shoubhik Debnath and Ronghang Hu and Didac Suris and Chaitanya Ryali and Kalyan Vasudev Alwala and Haitham Khedr and Andrew Huang and Jie Lei and Tengyu Ma and Baishan Guo and Arpit Kalla and Markus Marks and Joseph Greer and Meng Wang and Peize Sun and Roman Rädle and Triantafyllos Afouras and Effrosyni Mavroudi and Katherine Xu and Tsung-Han Wu and Yu Zhou and Liliane Momeni and Rishi Hazra and Shuangrui Ding and Sagar Vaze and Francois Porcher and Feng Li and Siyuan Li and Aishwarya Kamath and Ho Kei Cheng and Piotr Dollár and Nikhila Ravi and Kate Saenko and Pengchuan Zhang and Christoph Feichtenhofer},
      year={2025},
      eprint={2511.16719},
      archivePrefix={arXiv},
      primaryClass={cs.CV},
      url={https://arxiv.org/abs/2511.16719},
}

@inproceedings{chen2024elemental,
  title={ELEMENTAL: Interactive Learning from Demonstrations and Vision-Language Models for Reward Design in Robotics},
  author={Chen, Letian and Moorman, Nina Marie and Gombolay, Matthew Craig},
  year={2025},
  booktitle={Forty-second International Conference on Machine Learning}
}

@article{chi2024diffusionpolicy,
	author = {Cheng Chi and Zhenjia Xu and Siyuan Feng and Eric Cousineau and Yilun Du and Benjamin Burchfiel and Russ Tedrake and Shuran Song},
	title ={Diffusion Policy: Visuomotor Policy Learning via Action Diffusion},
	journal = {The International Journal of Robotics Research},
	year = {2024},
}

@article{chi2024universal,
  title={Universal manipulation interface: In-the-wild robot teaching without in-the-wild robots},
  author={Chi, Cheng and Xu, Zhenjia and Pan, Chuer and Cousineau, Eric and Burchfiel, Benjamin and Feng, Siyuan and Tedrake, Russ and Song, Shuran},
  journal={arXiv preprint arXiv:2402.10329},
  year={2024}
}

@inproceedings{dai2025racer,
  title={Racer: Rich language-guided failure recovery policies for imitation learning},
  author={Dai, Yinpei and Lee, Jayjun and Fazeli, Nima and Chai, Joyce},
  booktitle={2025 IEEE International Conference on Robotics and Automation (ICRA)},
  pages={15657--15664},
  year={2025},
  organization={IEEE}
}

@article{de2019causal,
  title={Causal confusion in imitation learning},
  author={De Haan, Pim and Jayaraman, Dinesh and Levine, Sergey},
  journal={Advances in neural information processing systems},
  volume={32},
  year={2019}
}

@inproceedings{dong2022webrobot,
  title={WebRobot: web robotic process automation using interactive programming-by-demonstration},
  author={Dong, Rui and Huang, Zhicheng and Lam, Ian Iong and Chen, Yan and Wang, Xinyu},
  booktitle={Proceedings of the 43rd ACM SIGPLAN International Conference on Programming Language Design and Implementation},
  pages={152--167},
  year={2022}
}

@inproceedings{dong-etal-2024-survey,
    title = "A Survey on In-context Learning",
    author = "Dong, Qingxiu  and
      Li, Lei  and
      Dai, Damai  and
      Zheng, Ce  and
      Ma, Jingyuan  and
      Li, Rui  and
      Xia, Heming  and
      Xu, Jingjing  and
      Wu, Zhiyong  and
      Chang, Baobao  and
      Sun, Xu  and
      Li, Lei  and
      Sui, Zhifang",
    editor = "Al-Onaizan, Yaser  and
      Bansal, Mohit  and
      Chen, Yun-Nung",
    booktitle = "Proceedings of the 2024 Conference on Empirical Methods in Natural Language Processing",
    month = nov,
    year = "2024",
    address = "Miami, Florida, USA",
    publisher = "Association for Computational Linguistics",
    url = "https://aclanthology.org/2024.emnlp-main.64/",
    doi = "10.18653/v1/2024.emnlp-main.64",
    pages = "1107--1128"
}

@article{driess2023palm,
  title={Palm-e: An embodied multimodal language model},
  author={Driess, Danny and Xia, Fei and Sajjadi, Mehdi SM and Lynch, Corey and Chowdhery, Aakanksha and Wahid, Ayzaan and Tompson, Jonathan and Vuong, Quan and Yu, Tianhe and Huang, Wenlong and others},
  year={2023}
}

@inproceedings{faulkner2023using,
  title={Using learning curve predictions to learn from incorrect feedback},
  author={Faulkner, Taylor A Kessler and Thomaz, Andrea L},
  booktitle={2023 IEEE International Conference on Robotics and Automation (ICRA)},
  pages={9414--9420},
  year={2023},
  organization={IEEE}
}

@inproceedings{fitzgerald2022inquire,
  title={Inquire: Interactive querying for user-aware informative reasoning},
  author={Fitzgerald, Tesca and Koppol, Pallavi and Callaghan, Patrick and Wong, Russell Quinlan Jun Hei and Simmons, Reid and Kroemer, Oliver and Admoni, Henny},
  booktitle={6th Annual Conference on Robot Learning},
  year={2022}
}

@article{fu2024mobile,
  title={Mobile aloha: Learning bimanual mobile manipulation with low-cost whole-body teleoperation},
  author={Fu, Zipeng and Zhao, Tony Z and Finn, Chelsea},
  journal={arXiv preprint arXiv:2401.02117},
  year={2024}
}

@article{gan2024can,
  title={Can Large Language Models Help Developers with Robotic Finite State Machine Modification?},
  author={Gan, Xiangyu Robin and Song, Yuxin Ray and Walker, Nick and Cakmak, Maya},
  journal={arXiv preprint arXiv:2412.05625},
  year={2024}
}

@article{golchha2024language,
  title={Language guided exploration for rl agents in text environments},
  author={Golchha, Hitesh and Yerawar, Sahil and Patel, Dhruvesh and Dan, Soham and Murugesan, Keerthiram},
  journal={arXiv preprint arXiv:2403.03141},
  year={2024}
}

@book{haykin1994neural,
  title={Neural networks: a comprehensive foundation},
  author={Haykin, Simon},
  year={1994},
  publisher={Prentice hall PTR}
}

@article{hu2025rac,
  title={RaC: Robot Learning for Long-Horizon Tasks by Scaling Recovery and Correction},
  author={Hu, Zheyuan and Wu, Robyn and Enock, Naveen and Li, Jasmine and Kadakia, Riya and Erickson, Zackory and Kumar, Aviral},
  journal={arXiv preprint arXiv:2509.07953},
  year={2025}
}

@inproceedings{huang2023anchor3dlane,
  title={Anchor3dlane: Learning to regress 3d anchors for monocular 3d lane detection},
  author={Huang, Shaofei and Shen, Zhenwei and Huang, Zehao and Ding, Zi-han and Dai, Jiao and Han, Jizhong and Wang, Naiyan and Liu, Si},
  booktitle={Proceedings of the IEEE/CVF Conference on Computer Vision and Pattern Recognition},
  pages={17451--17460},
  year={2023}
}

@inproceedings{hussenot2021hyperparameter,
  title={Hyperparameter selection for imitation learning},
  author={Hussenot, L{\'e}onard and Andrychowicz, Marcin and Vincent, Damien and Dadashi, Robert and Raichuk, Anton and Ramos, Sabela and Momchev, Nikola and Girgin, Sertan and Marinier, Raphael and Stafiniak, Lukasz and others},
  booktitle={International Conference on Machine Learning},
  pages={4511--4522},
  year={2021},
  organization={PMLR}
}

@inproceedings{kessler2021interactive,
  title={Interactive reinforcement learning from imperfect teachers},
  author={Kessler Faulkner, Taylor A and Thomaz, Andrea},
  booktitle={Companion of the 2021 ACM/IEEE international conference on human-robot interaction},
  pages={577--579},
  year={2021}
}

@misc{kim2025interactiveprogramsynthesismodeling,
      title={Interactive Program Synthesis for Modeling Collaborative Physical Activities from Narrated Demonstrations}, 
      author={Edward Kim and Daniel He and Jorge Chao and Wiktor Rajca and Mohammed Amin and Nishant Malpani and Ruta Desai and Antti Oulasvirta and Bjoern Hartmann and Sanjit Seshia},
      year={2025},
      eprint={2509.24250},
      archivePrefix={arXiv},
      primaryClass={cs.AI},
      url={https://arxiv.org/abs/2509.24250}, 
}

@article{kingma2014adam,
  title={Adam: A method for stochastic optimization},
  author={Kingma, Diederik P and Ba, Jimmy},
  journal={arXiv preprint arXiv:1412.6980},
  year={2014}
}

@article{laird2017interactive,
  title={Interactive task learning},
  author={Laird, John E and Gluck, Kevin and Anderson, John and Forbus, Kenneth D and Jenkins, Odest Chadwicke and Lebiere, Christian and Salvucci, Dario and Scheutz, Matthias and Thomaz, Andrea and Trafton, Greg and others},
  journal={IEEE Intelligent Systems},
  volume={32},
  number={4},
  pages={6--21},
  year={2017},
  publisher={IEEE}
}

@article{lee2025improving,
  title={Improving the Transparency of Robot Policies Using Demonstrations and Reward Communication},
  author={Lee, Michael S and Simmons, Reid and Admoni, Henny},
  journal={ACM Transactions on Human-Robot Interaction},
  volume={14},
  number={4},
  pages={1--31},
  year={2025},
  publisher={ACM New York, NY}
}

@inproceedings{li2025R,
  title={R*: Efficient Reward Design via Reward Structure Evolution and Parameter Alignment Optimization with Large Language Models},
  author={Li, Pengyi and Jianye, HAO and Tang, Hongyao and Yuan, Yifu and Qiao, Jinbin and Dong, Zibin and Zheng, Yan},
  booktitle={Forty-second International Conference on Machine Learning},
  year={2025}
}

@article{liang2022code,
  title={Code as policies: Language model programs for embodied control},
  author={Liang, Jacky and Huang, Wenlong and Xia, Fei and Xu, Peng and Hausman, Karol and Ichter, Brian and Florence, Pete and Zeng, Andy},
  journal={arXiv preprint arXiv:2209.07753},
  year={2022}
}

@article{lindstrom1988newton,
  title={Newton—Raphson and EM algorithms for linear mixed-effects models for repeated-measures data},
  author={Lindstrom, Mary J and Bates, Douglas M},
  journal={Journal of the American Statistical Association},
  volume={83},
  number={404},
  pages={1014--1022},
  year={1988},
  publisher={Taylor \& Francis}
}

@article{liu2024incomplete,
  title={An incomplete loop: Instruction inference, instruction following, and in-context learning in language models},
  author={Liu, Emmy and Neubig, Graham and Andreas, Jacob},
  journal={arXiv preprint arXiv:2404.03028},
  year={2024}
}

@article{liu2025factr,
  title={Factr: Force-attending curriculum training for contact-rich policy learning},
  author={Liu, Jason Jingzhou and Li, Yulong and Shaw, Kenneth and Tao, Tony and Salakhutdinov, Ruslan and Pathak, Deepak},
  journal={arXiv preprint arXiv:2502.17432},
  year={2025}
}

@inproceedings{luo2025human,
  title={Human-agent joint learning for efficient robot manipulation skill acquisition},
  author={Luo, Shengcheng and Peng, Quanquan and Lv, Jun and Hong, Kaiwen and Driggs--Campbell, Katherine Rose and Lu, Cewu and Li, Yong--Lu},
  booktitle={2025 IEEE International Conference on Robotics and Automation (ICRA)},
  pages={1370--1377},
  year={2025},
  organization={IEEE}
}

@misc{lynch2021languageconditionedimitationlearning,
      title={Language Conditioned Imitation Learning over Unstructured Data}, 
      author={Corey Lynch and Pierre Sermanet},
      year={2021},
      eprint={2005.07648},
      archivePrefix={arXiv},
      primaryClass={cs.RO},
      url={https://arxiv.org/abs/2005.07648}, 
}

@article{ma2023eureka,
  title={Eureka: Human-level reward design via coding large language models},
  author={Ma, Yecheng Jason and Liang, William and Wang, Guanzhi and Huang, De-An and Bastani, Osbert and Jayaraman, Dinesh and Zhu, Yuke and Fan, Linxi and Anandkumar, Anima},
  journal={arXiv preprint arXiv:2310.12931},
  year={2023}
}

@article{mao2023pdsketch,
  title={Pdsketch: Integrated planning domain programming and learning},
  author={Mao, Jiayuan and Lozano-P{\'e}rez, Tom{\'a}s and Tenenbaum, Joshua B and Kaelbling, Leslie Pack},
  journal={arXiv preprint arXiv:2303.05501},
  year={2023}
}

@inproceedings{nanavati2025lessons,
  title={Lessons Learned from Designing and Evaluating a Robot-Assisted Feeding System for Out-of-Lab Use},
  author={Nanavati, Amal and Gordon, Ethan K and Faulkner, Taylor A Kessler and Song, Yuxin Ray and Ko, Jonathan and Schrenk, Tyler and Nguyen, Vy and Zhu, Bernie Hao and Bolotski, Haya and Kashyap, Atharva and others},
  booktitle={2025 20th ACM/IEEE International Conference on Human-Robot Interaction (HRI)},
  pages={696--707},
  year={2025},
  organization={IEEE}
}

@inproceedings{o2024open,
  title={Open x-embodiment: Robotic learning datasets and rt-x models: Open x-embodiment collaboration 0},
  author={O’Neill, Abby and Rehman, Abdul and Maddukuri, Abhiram and Gupta, Abhishek and Padalkar, Abhishek and Lee, Abraham and Pooley, Acorn and Gupta, Agrim and Mandlekar, Ajay and Jain, Ajinkya and others},
  booktitle={2024 IEEE International Conference on Robotics and Automation (ICRA)},
  pages={6892--6903},
  year={2024},
  organization={IEEE}
}

@article{osa2018algorithmic,
  title={An algorithmic perspective on imitation learning},
  author={Osa, Takayuki and Pajarinen, Joni and Neumann, Gerhard and Bagnell, J Andrew and Abbeel, Pieter and Peters, Jan and others},
  journal={Foundations and Trends{\textregistered} in Robotics},
  volume={7},
  number={1-2},
  pages={1--179},
  year={2018},
  publisher={Now Publishers, Inc.}
}

@article{paszke2019pytorch,
  title={Pytorch: An imperative style, high-performance deep learning library},
  author={Paszke, Adam and Gross, Sam and Massa, Francisco and Lerer, Adam and Bradbury, James and Chanan, Gregory and Killeen, Trevor and Lin, Zeming and Gimelshein, Natalia and Antiga, Luca and others},
  journal={Advances in neural information processing systems},
  volume={32},
  year={2019}
}

@article{patton2024programming,
  title={Programming-by-demonstration for long-horizon robot tasks},
  author={Patton, Noah and Rahmani, Kia and Missula, Meghana and Biswas, Joydeep and Dillig, I{\c{s}}{\i}l},
  journal={Proceedings of the ACM on Programming Languages},
  volume={8},
  number={POPL},
  pages={512--545},
  year={2024},
  publisher={ACM New York, NY, USA}
}

@article{peng2024pragmatic,
  title={Pragmatic feature preferences: learning reward-relevant preferences from human input},
  author={Peng, Andi and Sun, Yuying and Shu, Tianmin and Abel, David},
  journal={arXiv preprint arXiv:2405.14769},
  year={2024}
}

@article{pomerleau1988alvinn,
  title={Alvinn: An autonomous land vehicle in a neural network},
  author={Pomerleau, Dean A},
  journal={Advances in neural information processing systems},
  volume={1},
  year={1988}
}

@inproceedings{ross2011reduction,
  title={A reduction of imitation learning and structured prediction to no-regret online learning},
  author={Ross, St{\'e}phane and Gordon, Geoffrey and Bagnell, Drew},
  booktitle={Proceedings of the fourteenth international conference on artificial intelligence and statistics},
  pages={627--635},
  year={2011},
  organization={JMLR Workshop and Conference Proceedings}
}

@article{salinas2024butterfly,
  title={The butterfly effect of altering prompts: How small changes and jailbreaks affect large language model performance},
  author={Salinas, Abel and Morstatter, Fred},
  journal={arXiv preprint arXiv:2401.03729},
  year={2024}
}

@article{saxena2025matters,
  title={What matters in learning from large-scale datasets for robot manipulation},
  author={Saxena, Vaibhav and Bronars, Matthew and Arachchige, Nadun Ranawaka and Wang, Kuancheng and Shin, Woo Chul and Nasiriany, Soroush and Mandlekar, Ajay and Xu, Danfei},
  journal={arXiv preprint arXiv:2506.13536},
  year={2025}
}

@article{schuirmann1987comparison,
  title={A comparison of the two one-sided tests procedure and the power approach for assessing the equivalence of average bioavailability},
  author={Schuirmann, Donald J},
  journal={Journal of pharmacokinetics and biopharmaceutics},
  volume={15},
  number={6},
  pages={657--680},
  year={1987},
  publisher={Springer}
}

@article{shao2019mental,
  title={Mental workload characteristics of manipulator teleoperators with different spatial cognitive abilities},
  author={Shao, Shuyu and Zhou, Qianxiang and Liu, Zhongqi},
  journal={International journal of advanced robotic systems},
  volume={16},
  number={6},
  pages={1729881419888042},
  year={2019},
  publisher={SAGE Publications Sage UK: London, England}
}

@article{shi2024yell,
  title={Yell at your robot: Improving on-the-fly from language corrections},
  author={Shi, Lucy Xiaoyang and Hu, Zheyuan and Zhao, Tony Z and Sharma, Archit and Pertsch, Karl and Luo, Jianlan and Levine, Sergey and Finn, Chelsea},
  journal={arXiv preprint arXiv:2403.12910},
  year={2024}
}

@article{tang2023saytap,
  title={Saytap: Language to quadrupedal locomotion},
  author={Tang, Yujin and Yu, Wenhao and Tan, Jie and Zen, Heiga and Faust, Aleksandra and Harada, Tatsuya},
  journal={arXiv preprint arXiv:2306.07580},
  year={2023}
}

@article{towers2024gymnasium,
  title={Gymnasium: A standard interface for reinforcement learning environments},
  author={Towers, Mark and Kwiatkowski, Ariel and Terry, Jordan and Balis, John U and De Cola, Gianluca and Deleu, Tristan and Goul{\~a}o, Manuel and Kallinteris, Andreas and Krimmel, Markus and KG, Arjun and others},
  journal={arXiv preprint arXiv:2407.17032},
  year={2024}
}

@article{vsidak1967rectangular,
  title={Rectangular confidence regions for the means of multivariate normal distributions},
  author={{\v{S}}id{\'a}k, Zbyn{\v{e}}k},
  journal={Journal of the American statistical association},
  volume={62},
  number={318},
  pages={626--633},
  year={1967},
  publisher={Taylor \& Francis}
}

@article{wald1943tests,
  title={Tests of statistical hypotheses concerning several parameters when the number of observations is large},
  author={Wald, Abraham},
  journal={Transactions of the American Mathematical society},
  volume={54},
  number={3},
  pages={426--482},
  year={1943},
  publisher={JSTOR}
}

@article{wang2024rl,
  title={Rl-vlm-f: Reinforcement learning from vision language foundation model feedback},
  author={Wang, Yufei and Sun, Zhanyi and Zhang, Jesse and Xian, Zhou and Biyik, Erdem and Held, David and Erickson, Zackory},
  journal={arXiv preprint arXiv:2402.03681},
  year={2024}
}

@article{wang2025cori,
  title={CoRI: Synthesizing Communication of Robot Intent for Physical Human-Robot Interaction},
  author={Wang, Junxiang and K{\"u}{\c{c}}{\"u}ktabak, Emek Bar{\i}{\c{s}} and Zarrin, Rana Soltani and Erickson, Zackory},
  journal={arXiv preprint arXiv:2505.20537},
  year={2025}
}

@article{wei2022chain,
  title={Chain-of-thought prompting elicits reasoning in large language models},
  author={Wei, Jason and Wang, Xuezhi and Schuurmans, Dale and Bosma, Maarten and Xia, Fei and Chi, Ed and Le, Quoc V and Zhou, Denny and others},
  journal={Advances in neural information processing systems},
  volume={35},
  pages={24824--24837},
  year={2022}
}

@article{wei2022emergent,
  title={Emergent abilities of large language models},
  author={Wei, Jason and Tay, Yi and Bommasani, Rishi and Raffel, Colin and Zoph, Barret and Borgeaud, Sebastian and Yogatama, Dani and Bosma, Maarten and Zhou, Denny and Metzler, Donald and others},
  journal={arXiv preprint arXiv:2206.07682},
  year={2022}
}

@article{welch1938significance,
  title={The significance of the difference between two means when the population variances are unequal},
  author={Welch, Bernard L},
  journal={Biometrika},
  volume={29},
  number={3/4},
  pages={350--362},
  year={1938},
  publisher={JSTOR}
}

@inproceedings{wu2019imitation,
  title={Imitation learning from imperfect demonstration},
  author={Wu, Yueh-Hua and Charoenphakdee, Nontawat and Bao, Han and Tangkaratt, Voot and Sugiyama, Masashi},
  booktitle={International Conference on Machine Learning},
  pages={6818--6827},
  year={2019},
  organization={PMLR}
}

@inproceedings{wu2024gello,
  title={Gello: A general, low-cost, and intuitive teleoperation framework for robot manipulators},
  author={Wu, Philipp and Shentu, Yide and Yi, Zhongke and Lin, Xingyu and Abbeel, Pieter},
  booktitle={2024 IEEE/RSJ International Conference on Intelligent Robots and Systems (IROS)},
  pages={12156--12163},
  year={2024},
  organization={IEEE}
}

@article{yu2022using,
  title={Using both demonstrations and language instructions to efficiently learn robotic tasks},
  author={Yu, Albert and Mooney, Raymond J},
  journal={arXiv preprint arXiv:2210.04476},
  year={2022}
}

@inproceedings{zhao2022supporting,
  title={Supporting End Users in Defining Reinforcement-Learning Problems for Human-Robot Interactions},
  author={Zhao, Valerie and Littman, Michael L and Lu, Shan and Sebo, Sarah and Ur, Blase},
  booktitle={The 5th Multidisciplinary Conference on Reinforcement Learning and Decision Making (RLDM)},
  year={2022}
}

@article{zhao2023learning,
  title={Learning fine-grained bimanual manipulation with low-cost hardware},
  author={Zhao, Tony Z and Kumar, Vikash and Levine, Sergey and Finn, Chelsea},
  journal={arXiv preprint arXiv:2304.13705},
  year={2023}
}

@article{zhao2024aloha,
  title={Aloha unleashed: A simple recipe for robot dexterity},
  author={Zhao, Tony Z and Tompson, Jonathan and Driess, Danny and Florence, Pete and Ghasemipour, Kamyar and Finn, Chelsea and Wahid, Ayzaan},
  journal={arXiv preprint arXiv:2410.13126},
  year={2024}
}

@article{zhu2023ghost,
  title={Ghost in the minecraft: Generally capable agents for open-world environments via large language models with text-based knowledge and memory},
  author={Zhu, Xizhou and Chen, Yuntao and Tian, Hao and Tao, Chenxin and Su, Weijie and Yang, Chenyu and Huang, Gao and Li, Bin and Lu, Lewei and Wang, Xiaogang and others},
  journal={arXiv preprint arXiv:2305.17144},
  year={2023}
}

@inproceedings{zhu2024bootstrapping,
  title={Bootstrapping cognitive agents with a large language model},
  author={Zhu, Feiyu and Simmons, Reid},
  booktitle={Proceedings of the AAAI Conference on Artificial Intelligence},
  volume={38},
  number={1},
  pages={655--663},
  year={2024}
}

@article{zhu2025sample,
  title={Sample-Efficient Behavior Cloning Using General Domain Knowledge},
  author={Zhu, Feiyu and Oh, Jean and Simmons, Reid},
  journal={arXiv preprint arXiv:2501.16546},
  year={2025}
}

@String{Computer = "{IEEE} Computer" }

@String{Chelsea = "Chelsea" }

@String{Springer = "Springer-Verlag" }

% % %%
% % %% If your work has an appendix, this is the place to put it.
% % 
\clearpage
\appendix
\onecolumn

\section{Interface Design}

\begin{figure}[h!]
    \centering
    \includegraphics[width=0.75\linewidth]{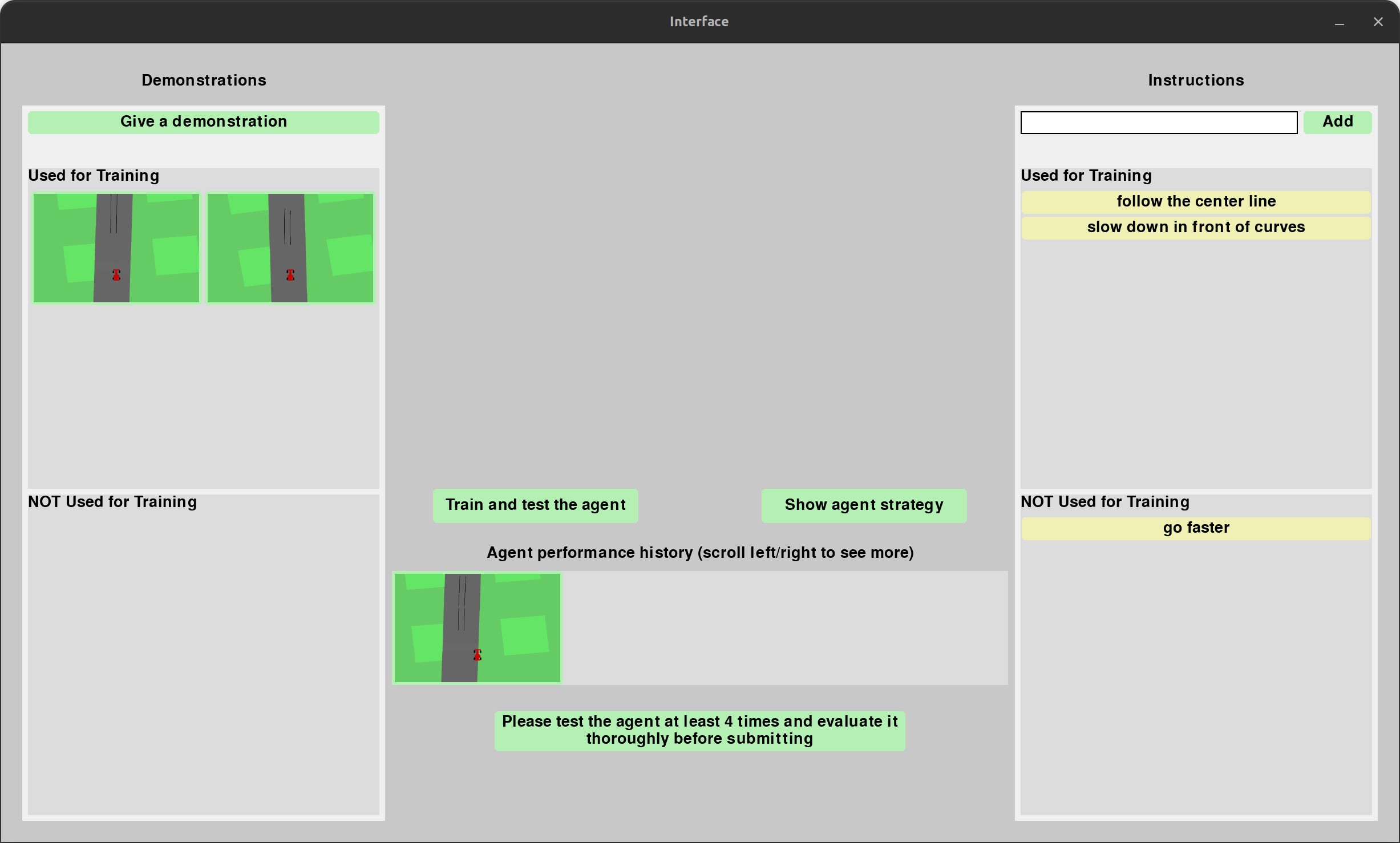}
    \caption{Home page of the interface}
    \label{fig:interface-home}
    \Description{
    The home page of the interaction. It shows three columns.
    On the left is a demonstration panel where there is a "give a demonstration" button on the top, followed by two box each saying "used in training" and "not used in training".
    For the box that says "used in training" there are two thumbnails each showing a red race car in the middle of a race track. The box that says "not used for training" is empty.
    In the lower middle part of the interface are two buttons that each says "train and test the agent" and "show agent strategy".
    Moving down there is a section titled "agent performance history (scroll left/right to see more)".
    Moving down there is another thumbnail of a red race car on a track surrounded by grass.
    At the button of the interface is a wide button that says "please test the agent at least 4 times and evaluate it thoroughly before submitting".
    On the right there is an instruction panel that is very similar to the demonstration panel.
    But instead of the "give a demonstration" button there is a text box with an "add" button next to it.
    There are two slices of instructions that are in the used for training box and they read "follow the center line", "slow down in front of curves". The box that is not used for training has one slice of instruction that says "go faster".
    }
\end{figure}

\begin{figure}[h!]
    \centering
    \includegraphics[width=0.75\linewidth]{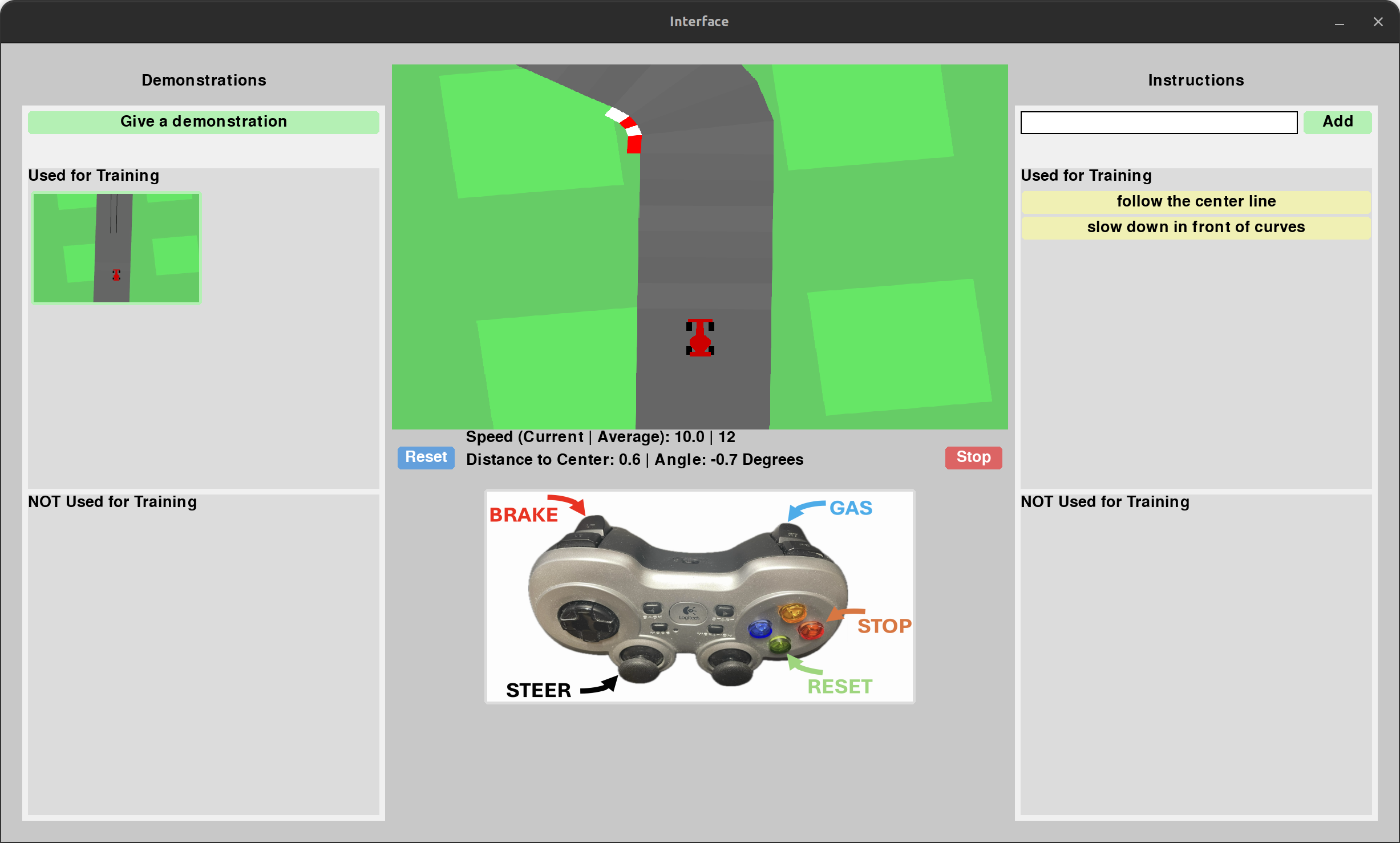}
    \caption{Demonstration page of the interface}
    \label{fig:interface-demo}
    \Description{
    The home page of the interaction. It shows three columns.
    On the left is a demonstration panel where there is a "give a demonstration" button on the top, followed by two box each saying "used in training" and "not used in training".
    For the box that says "used in training" there are one thumbnail showing a red race car in the middle of a race track. The box that says "not used for training" is empty.
    In the middle it shows a big screen of a red race car on a track. Below the screen there is a blue reset button on the left and a red top button on the right. In the middle it shows the current speed, average speed, distance to center, and angle information of the car.
    Below them is an illustration of how to use the game pad controller.
    Specifically, the left and right triggers are used for brake and gas respectively, the left joystick is used for steering, and the A/B buttons are used for reset and stop respectively.
    On the right there is an instruction panel that is very similar to the demonstration panel.
    But instead of the "give a demonstration" button there is a text box with an "add" button next to it.
    There are two slices of instructions that are in the used for training box and they read "follow the center line", "slow down in front of curves". The box that is not used for training is empty.
    }
\end{figure}

Figures \ref{fig:interface-home} and \ref{fig:interface-demo} show the home page and demonstration page of the interface, respectively.
On the home page, the user can choose to give a demonstration, give a new instruction, move a demonstration / instruction to not be used for training and back, train and test the agent, review its strategy in natural language, and review the previous demonstrations and rollouts.
For the baseline condition, the instruction panel on the right and the ``show agent strategy" button are unavailable, but the rest is identical.


\newpage
\section{Prompts Used}

System prompt $\mathcal{S}$ used in prompting:

\tagpdfsetup{para/tagging=false}
\begin{lstlisting}[breaklines=true, numbers=left, frame=single]
Implement pytorch models that reflect the specified structure of the user.

The user will provide the following:
* [High-Level Instructions] explains the decision-making principles of the model
* [Features] explains how to interpret the input to the model. For example, which dimension of the input corresponds to which feature, and what type (discrete or continuous)
* [Output Space] explains the action space
* [Additional Notes] (optional) explains any additional details of the task

You will do the following steps before giving the final implementation:

* [Variables Extraction] List the names of the variables, including feature space, latent space, and output space. Note that some of them need to be inferred from the instructions.

* [Structure Description] Describe the connections / operations between each variables / features.

* [Plan the connections] List the variables (and their type and shape) in the order in which they should be computed, where the variables based only on the input features are listed first, then the variables that depend on those, etc. For each of them, explain how they can be computed using the previously listed variables or inputs to the model. Also, indicate whether the new feature is positively correlated to the previous feature or negatively correlated respectively. Be very specific. List the functions or operators that should be used to connect the variables. If you decided to use a linear combination, explain why a bias term is included or not included.

Use the following format:
[Variables]
 * Name of the first variable (shape and type)
 * Name of the second variable (shape and type)
 ...

[Structure Description]
English description of the model.

[Connections]
 * Name of the first variable that should be computed (shape and type)
    - depends on <feature 1 name> (positively correlated), <feature 2 name> (nagatively correlated), ...
    - can be computed with a linear combination of ... and with a bias term
    - the bias term is included because ...
 * Name of the second variable that should be computed (shape and type)
    - depends on <feature 1 name> (negatively correlated), <feature 2 name> (positively correlated), ...
    - can be computed using `torch.where` on ...

[Code]
```py
import torch
from torch import nn

class ModelName(nn.Module):
    <YOUR MODEL DEFINITION>
```


Notes:
* Represent new features as linear combinations of old features when possible. If you are very certain that no bias term is needed (i.e., the feature value should be 0 if all inputs are 0), then don't include the bias term to make it easier to learn.
* Register all weights and bias terms as `nn.Parameter` with `required_grad=True`. Don't use `register_buffer`.
* Don't name a parameter using pytorch keywords such as `self.half`.
* There might be cases where a linear combination is not sufficient; then you may use other operations, such as multiplication, to represent the interaction between two features.
* Keep the model simple.
* When initializing the values for each weight and bias, set a value based on positive or negative correlation (e.g., if the input feature is negatively correlated, then set its weight to -0.1) instead of using random initialization. Keep the initialized value between -0.5 and 0.5.
* There should be no magic number in the `forward` function. Constants like 0, 1, and 0.5 are fine.
* Make sure all operations are differentiable so that the parameters can be learned by gradient descent.
* You may use any of the functions defined in pytorch (e.g., `torch.logical_and`, `torch.clamp`, `torch.abs`, `torch.square`, etc.).
* Refrain from using `torch.sigmoid` or `torch.tanh` to avoid the vanishing gradient problem.
* Make sure the gradient can flow back to the parameters. Avoid in-place operations (always use a new variable name instead) or constructing new tensors (always use `torch.stack` or `torch.cat` instead).
* You may assume the inputs are already normalized.
* Assume the inputs are unbatched.
* For a discrete output space, the model should output a (potentially unnormalized) distribution among those discrete actions. For a continuous action space, return the predicted action without worrying about distributions.
\end{lstlisting}\par
\tagpdfsetup{para/tagging=true}

Few shot example $\mathcal{I}_\text{lander}$ used in prompting

\tagpdfsetup{para/tagging=false}
\begin{lstlisting}[breaklines=true, numbers=left, frame=single]
[High-Level Instruction]

 * The lander heading should always point to the center
 * Don't tilt the lander too much
 * If the lander is too low and too far from the landing pad, then it should activte the main engine
 * Use the main engine to slow the lander down if it's falling too fast
 * When then lander contacts the ground, only use the main engine to slow down
 * Don't active any engine if there is no need to

[Features]
The input to the model is a tensor of $(8)$. The features of the lander in each dimension are:
0. (float32) horizontal coordinate $x$
1. (float32) vertical coordinate $y$
2. (float32) horizontal speed $v_x$
3. (float32) vertical speed $v_y$
4. (float32) heading $\theta$
5. (float32) angular velocity $\omega$
6. (bool) whether the left landing leg is in contact with the ground
7. (bool) whether the right landing leg is in contact with the ground
[Output Space]
A tensor of shape (4,) representing the unnormalized distribution among the following four actions
0: do nothing
1: fire left orientation engine
2: fire main engine
3: fire right orientation engine

[Additional Notes]

The lander is upright when $\theta = 0$ and is tilting to the left when $\theta > 0$. When the lander is falling $v_y < 0$ since the y-axis points upward.
The landing pad is always at (0, 0).
The left and right engines are semetric.
Name your model "LanderPolicy".
\end{lstlisting}\par
\tagpdfsetup{para/tagging=true}

Few shot example $\llbracket \alpha \rrbracket_\text{Lander}$ used in prompting

\tagpdfsetup{para/tagging=false}
\begin{lstlisting}[breaklines=true, numbers=left, frame=single]
[Variables]
 * x (float32, shape=())
 * y (float32, shape=())
 * v_x (float32, shape=())
 * v_y (float32, shape=())
 * theta (float32, shape=())
 * omega (float32, shape=())
 * left_leg_contact (bool, shape=())
 * right_leg_contact (bool, shape=())
 * in_air (bool, shape=()): True if both legs are not in contact with the ground
 * target_heading (float32, shape=()): desired heading, clipped to a range
 * target_y (float32, shape=()): desired vertical coordinate, depends on |x|
 * heading_adjustment (float32, shape=()): difference between current and target heading, and omega
 * speed_adjustment (float32, shape=()): proportional to v_y
 * vertical_adjustment (float32, shape=()): difference between current and target y, and v_y
 * left_engine_prob (float32, shape=()): probability of firing left engine
 * right_engine_prob (float32, shape=()): probability of firing right engine
 * main_engine_prob (float32, shape=()): probability of firing main engine
 * base_prob (float32, shape=()): base probability of doing nothing
 * output (float32, shape=(4,)): unnormalized logits for each action

[Structure Description]
 * The lander is in the air if none of its legs are in contact with the ground. Otherwise, it is in contact with the ground.
 * The target heading of the lander depends on its horizontal coordinate and speed so that it points to the center. But we will clip in a range such that it stays roughly in the middle, because tilting too much is bad.
 * The target vertical coordinate depends on the magnitude of the horizontal offset of the lander. The further the lander is from the landing pad (which is at $(0, 0)$), the higher the target vertical coordinate should be.
 * The heading adjustment depends on the difference between the current and clipped target heading of the lander, as well as the current angular velocity.
 * The speed adjustment needed to put the lander to rest is proportional to its vertical speed.
 * And the vertical adjustment depends on the difference between the current and target vertical coordinate, as well as the vertical speed.
 * Only activate the left or the right engine when the lander is not contacting the ground. And the probability of activating the left engine is the heading adjustment, and symmetrically, the probability of activating the right engine is the negation of the heading adjustment.
 * The probability of activating the main engine in the air is the vertical adjustment.
 * The probability of activating the main engine when the lander is in contact with the ground is the speed adjustment.
 * There is a base level probability that the lander will do nothing regardless of the input.

[Connections]
 * in_air (bool, shape=())
    - depends on left_leg_contact (negatively correlated), right_leg_contact (negatively correlated)
    - can be computed as `~(left_leg_contact | right_leg_contact)`
 * target_heading (float32, shape=())
    - depends on x (negatively correlated), v_x (negatively correlated)
    - can be computed as a linear combination: $w_1 x + w_2 v_x$ (both negative weights), then clipped to $[-\theta_{max}, \theta_{max}]$
    - bias term is not included because if x and v_x are 0, the target heading should be 0
 * target_y (float32, shape=())
    - depends on x (positively correlated)
    - can be computed as $w_3 |x|$ (positive weight), no bias term (if x=0, target_y=0)
 * heading_adjustment (float32, shape=())
    - depends on theta (negatively correlated), target_heading (positively correlated), omega (negatively correlated)
    - can be computed as $w_4 (target\_heading - theta) + w_5 (-omega)$, no bias term
 * speed_adjustment (float32, shape=())
    - depends on v_y (negatively correlated)
    - can be computed as $w_6 (-v_y)$, no bias term
 * vertical_adjustment (float32, shape=())
    - depends on y (negatively correlated), target_y (positively correlated), v_y (negatively correlated)
    - can be computed as $w_7 (target\_y - y) + w_8 (-v_y)$, no bias term
 * left_engine_prob (float32, shape=())
    - depends on in_air (positively correlated), heading_adjustment (positively correlated)
    - can be computed as $in\_air * heading\_adjustment$
 * right_engine_prob (float32, shape=())
    - depends on in_air (positively correlated), heading_adjustment (negatively correlated)
    - can be computed as $in\_air * (-heading\_adjustment)$
 * main_engine_prob (float32, shape=())
    - depends on in_air (positively correlated), vertical_adjustment (positively correlated), speed_adjustment (positively correlated)
    - can be computed as $in\_air * vertical\_adjustment + (1-in\_air) * speed\_adjustment$
 * base_prob (float32, shape=())
    - independent, learnable parameter
 * output (float32, shape=(4,))
    - stack [base_prob, left_engine_prob, main_engine_prob, right_engine_prob]

[Code]
```py
import torch
from torch import nn

class LanderPolicy(nn.Module):
    def __init__(self):
        super().__init__()
        # Parameters for target_heading: target_heading = w1 * x + w2 * v_x, clipped to [-theta_max, theta_max]
        self.w1 = nn.Parameter(torch.tensor(-0.5, dtype=torch.float32))  # negative correlation with x
        self.w2 = nn.Parameter(torch.tensor(-0.5, dtype=torch.float32))  # negative correlation with v_x
        self.theta_max = nn.Parameter(torch.tensor(0.4, dtype=torch.float32), requires_grad=False)  # non-gradient parameter

        # Parameter for target_y: target_y = w3 * |x|
        self.w3 = nn.Parameter(torch.tensor(0.5, dtype=torch.float32))  # positive correlation with |x|

        # Parameters for heading_adjustment: w4 * (target_heading - theta) + w5 * (-omega)
        self.w4 = nn.Parameter(torch.tensor(1.0, dtype=torch.float32))  # positive correlation with (target_heading - theta)
        self.w5 = nn.Parameter(torch.tensor(-0.5, dtype=torch.float32))  # negative correlation with omega

        # Parameter for speed_adjustment: w6 * (-v_y)
        self.w6 = nn.Parameter(torch.tensor(0.5, dtype=torch.float32))  # positive correlation with -v_y

        # Parameters for vertical_adjustment: w7 * (target_y - y) + w8 * (-v_y)
        self.w7 = nn.Parameter(torch.tensor(1.0, dtype=torch.float32))  # positive correlation with (target_y - y)
        self.w8 = nn.Parameter(torch.tensor(0.5, dtype=torch.float32))  # positive correlation with -v_y

        # Base probability for "do nothing"
        self.base_prob = nn.Parameter(torch.tensor(0.1, dtype=torch.float32))

    def forward(self, obs):
        # obs: shape (8,)
        x = obs[0]
        y = obs[1]
        v_x = obs[2]
        v_y = obs[3]
        theta = obs[4]
        omega = obs[5]
        left_leg_contact = obs[6].bool()
        right_leg_contact = obs[7].bool()

        # in_air: True if both legs are not in contact
        in_air = (~(left_leg_contact | right_leg_contact)).float()

        # target_heading = w1 * x + w2 * v_x, clipped to [-theta_max, theta_max]
        unclipped_target_heading = self.w1 * x + self.w2 * v_x
        target_heading = torch.clamp(unclipped_target_heading, -self.theta_max, self.theta_max)

        # target_y = w3 * |x|
        target_y = self.w3 * torch.abs(x)

        # heading_adjustment = w4 * (target_heading - theta) + w5 * (-omega)
        heading_adjustment = self.w4 * (target_heading - theta) + self.w5 * (-omega)

        # speed_adjustment = w6 * (-v_y)
        speed_adjustment = self.w6 * (-v_y)

        # vertical_adjustment = w7 * (target_y - y) + w8 * (-v_y)
        vertical_adjustment = self.w7 * (target_y - y) + self.w8 * (-v_y)

        # left_engine_prob = in_air * heading_adjustment
        left_engine_prob = in_air * heading_adjustment

        # right_engine_prob = in_air * (-heading_adjustment)
        right_engine_prob = in_air * (-heading_adjustment)

        # main_engine_prob = in_air * vertical_adjustment + (1-in_air) * speed_adjustment
        main_engine_prob = in_air * vertical_adjustment + (1.0 - in_air) * speed_adjustment

        # base_prob is a learnable parameter
        base_prob = self.base_prob

        # Output: [do nothing, left engine, main engine, right engine]
        output = torch.stack([base_prob, left_engine_prob, main_engine_prob, right_engine_prob], dim=0)
        return output
```
\end{lstlisting}\par
\tagpdfsetup{para/tagging=true}

User prompt that takes in $\mathcal{I}_i$
% Few shot example $\mathcal{I}_\text{lander}$ used in prompting

\tagpdfsetup{para/tagging=false}
\begin{lstlisting}[breaklines=true, frame=single, numbers=left]
[High-Level Instruction]
* [USER_INSTRUCTIONS]

[Features]
The input to the model is two tensors representing the tracks and other indicator information:

0. (float32, (B, L, 7)) tiles. Where $B$ is the batch size and $L=8$ is the number of tiles on the track ahead of the racecar, and the last dimension contains $(x, y, _, _, \theta, _, border)$ where $x$ is the signed lateral position of the tile (positive means the race car is on the left of the center of the tile), $y$ is the signed longitudinal position of the tile, $\theta$ is the relative heading of the tile to the racecar (positive means the road is heading to the left), and $border$ is either 0 or 1, indicating whether the tile is a sharp corner (curvature exceeds some threshold). Other features from input (marked by $_$) can be ignored.
1. (float32, (B, 7)) indicators. Where the first column is the speed $v$ (normalized from [0, 100] to [0, 1]) and the second column is the current (absolute) heading of the race car $\theta$ (normalized from [-pi, pi] to [-1, 1]), while the other features are not important.

[Output Space]
A tensor of shape (B, 3) where each element corresponds to steer, accelerate, and brake controls of the race car, respectively. These controls are continuous. The steer value should be in the range of $(-1, 1)$, where -1 is full left, 0 is no steering, and 1 is full right. Accelerate is in the range of $(0, 1)$, where 0 is no acceleration and 1 is full acceleration. Brake is also in the range of $(0, 1)$, where 0 is no braking and 1 is full braking.

[Additional Notes]
* The track is a loop.
* The track is about 0.4 unit wide. When the closest tile's x is at -0.2 or 0.2 it means the racecar is on the edge of the track. Beyond that it's grass.
* The racecar is approximately 3 tiles long.
* The first tile is the closest tile to the race car, and the rest are in the order of tiles ahead.
* The arguments of the forward function should only be self, tiles, and indicators.
* Clip the output to the desired range.
* Name your model "RacecarPolicy".

\end{lstlisting}\par
\tagpdfsetup{para/tagging=true}

\newpage
\section{Participant Background}

\begin{figure}[h]
    \centering
    \includegraphics[width=0.7\linewidth]{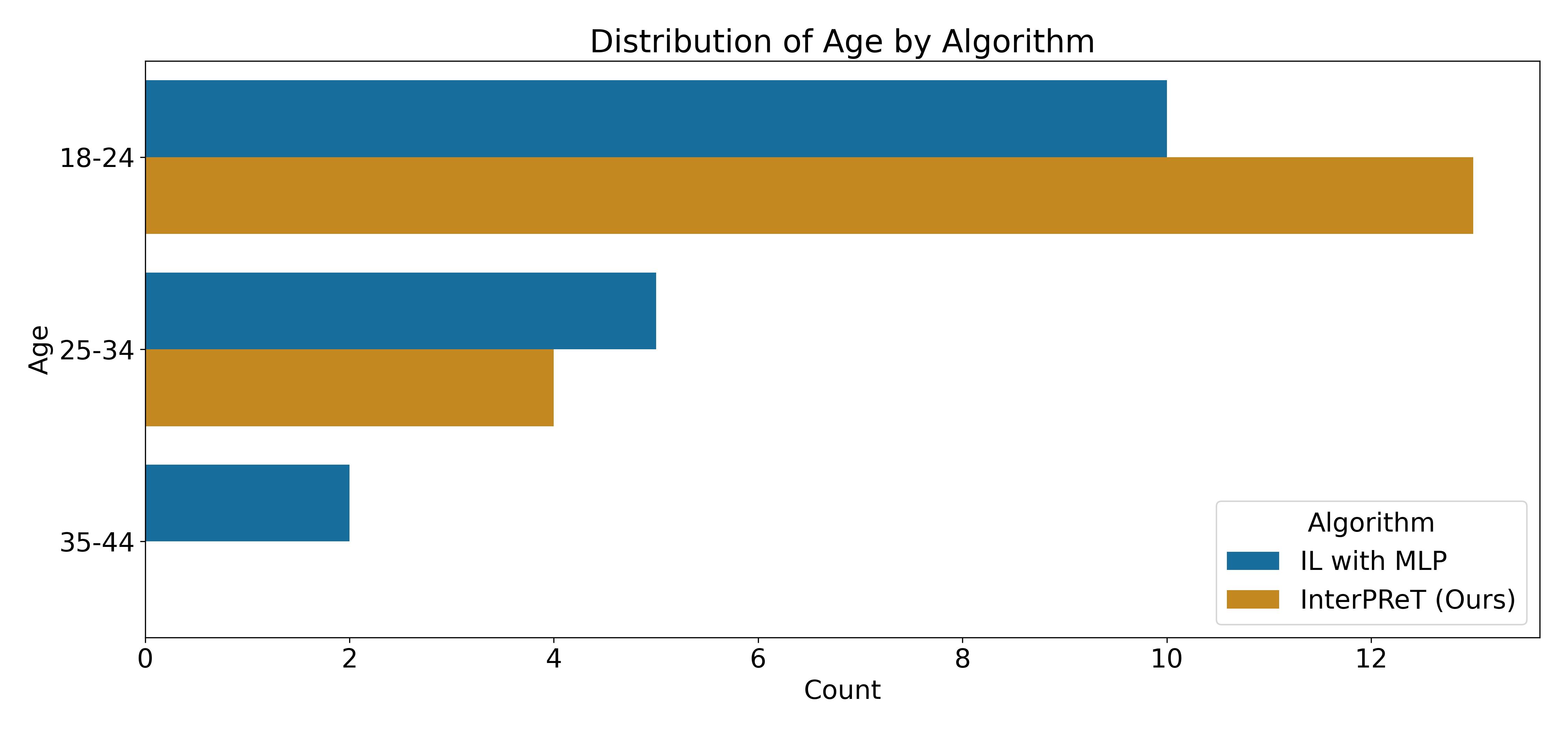}
    \caption{Age Distribution}
    \label{fig:count-age}
    \Description{
    A count plot that shows the age distribution. 
    Specifically, for the baseline condition, there are 10 participants of age 18 to 24, 5 participants of age 25 to 34, and 2 participants of age 35 to 44.
    For the interpret condition, there are 13 participants of age 18 to 24, 4 participants of age 25 to 34.
    }
\end{figure}

\begin{figure}[h]
    \centering
    \includegraphics[width=0.7\linewidth]{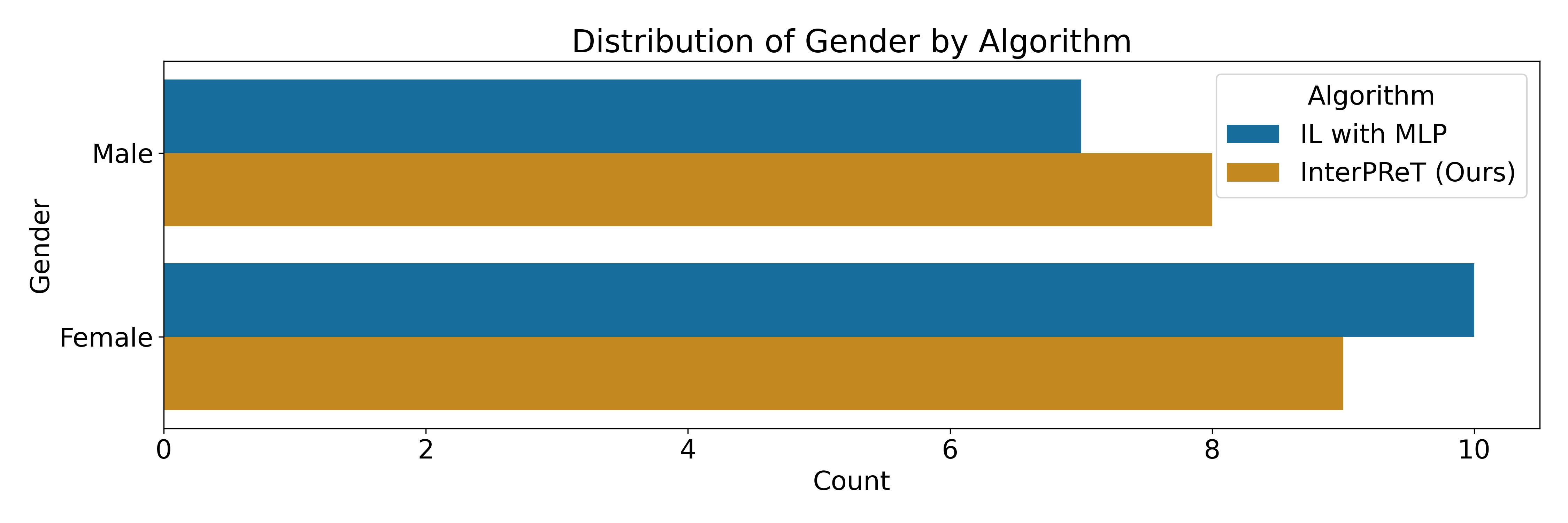}
    \caption{Gender Distribution}
    \label{fig:count-gender}
    \Description{
    A count plot that shows the gender distribution. 
    Specifically, for the baseline condition, there are 7 participants that identifies as males, and 10 that identifies as females.
    For the interpret condition, there are 8 participants that identifies as males, and 9 that identifies as females.
    }
\end{figure}

\begin{figure}[h]
    \centering
    \includegraphics[width=0.7\linewidth]{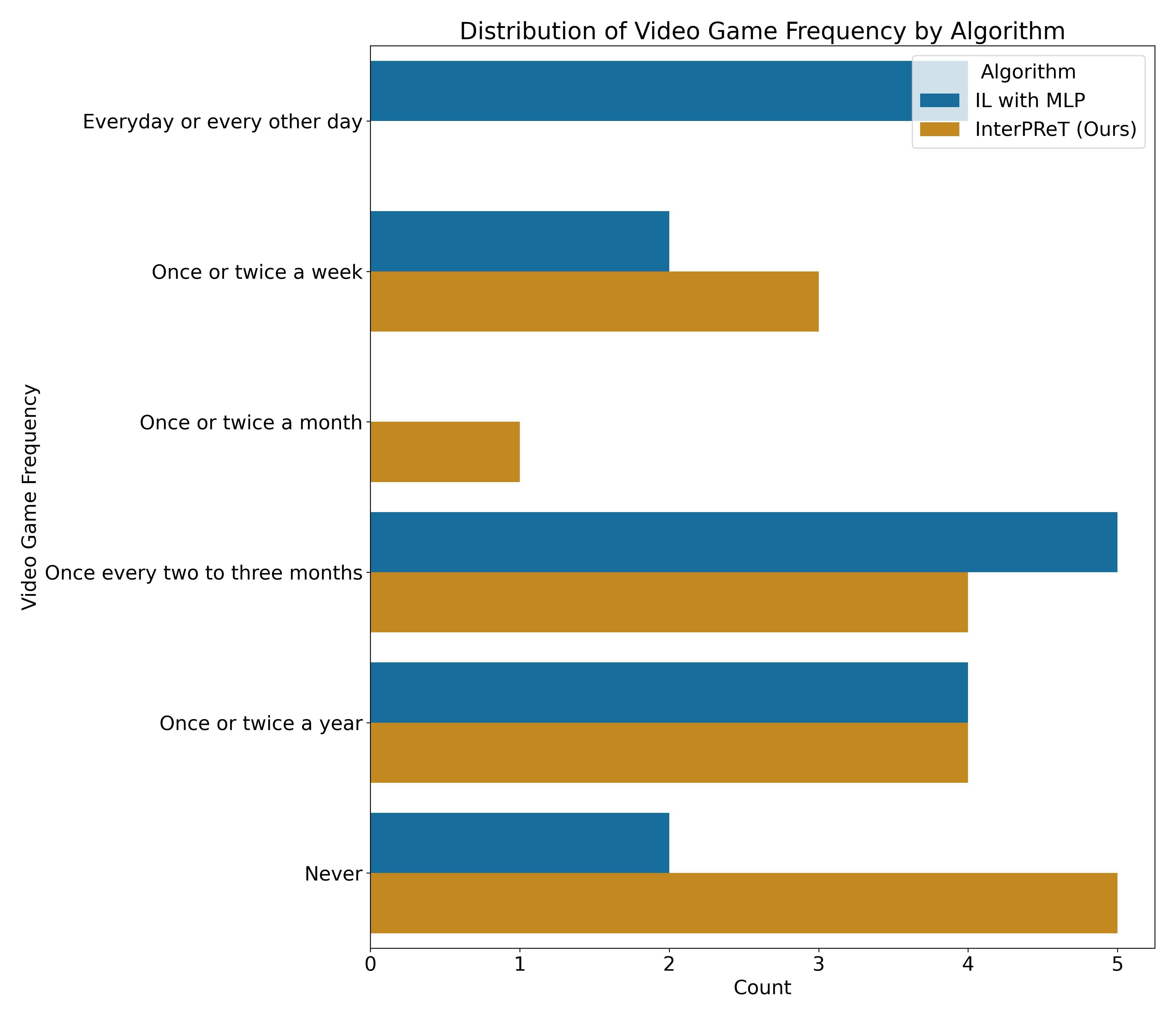}
    \caption{Game Frequency Distribution}
    \label{fig:count-game}
    \Description{
    A count plot that shows the distribution of video game experience.
    Specifically, for the baseline condition, there are 4 participants reported playing video games everyday or every other day, 2 for once or twice a week, 5 for once every two to three months, 4 for once or twice a year, and 2 for never.
    For the interpret condition, there are 3 participants reported playing video games for once or twice a week, 1 for once or twice a month, 4 for once every two to three months, 4 for once or twice a year, and 5 for never.
    }
\end{figure}

\begin{figure}[h]
    \centering
    \includegraphics[width=0.7\linewidth]{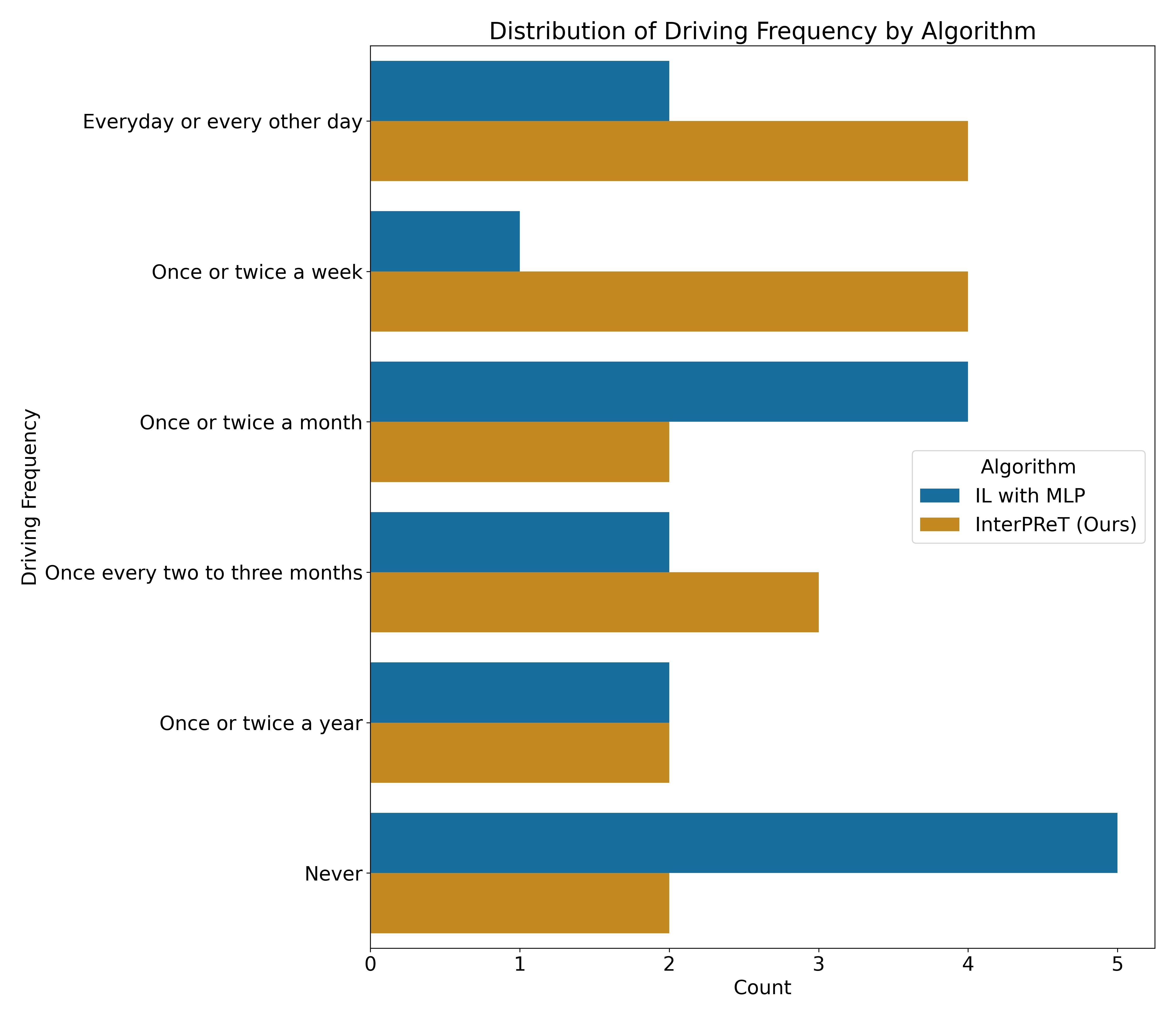}
    \caption{Driving Frequency Distribution}
    \label{fig:count-driving}
    \Description{
    A count plot that shows the distribution of driving experience.
    Specifically, for the baseline condition, there are 2 participants reported driving everyday or every other day, 1 for once or twice a week, 4 for once or twice a week, 2 for once every two to three months, 2 for once or twice a year, and 5 for never.
    For the interpret condition, there are 4 participants reported driving everyday or every other day, 4 for once or twice a week, 2 for once or twice a week, 3 for once every two to three months, 2 for once or twice a year, and 2 for never.
    }
\end{figure}

\begin{figure}[h]
    \centering
    \includegraphics[width=0.7\linewidth]{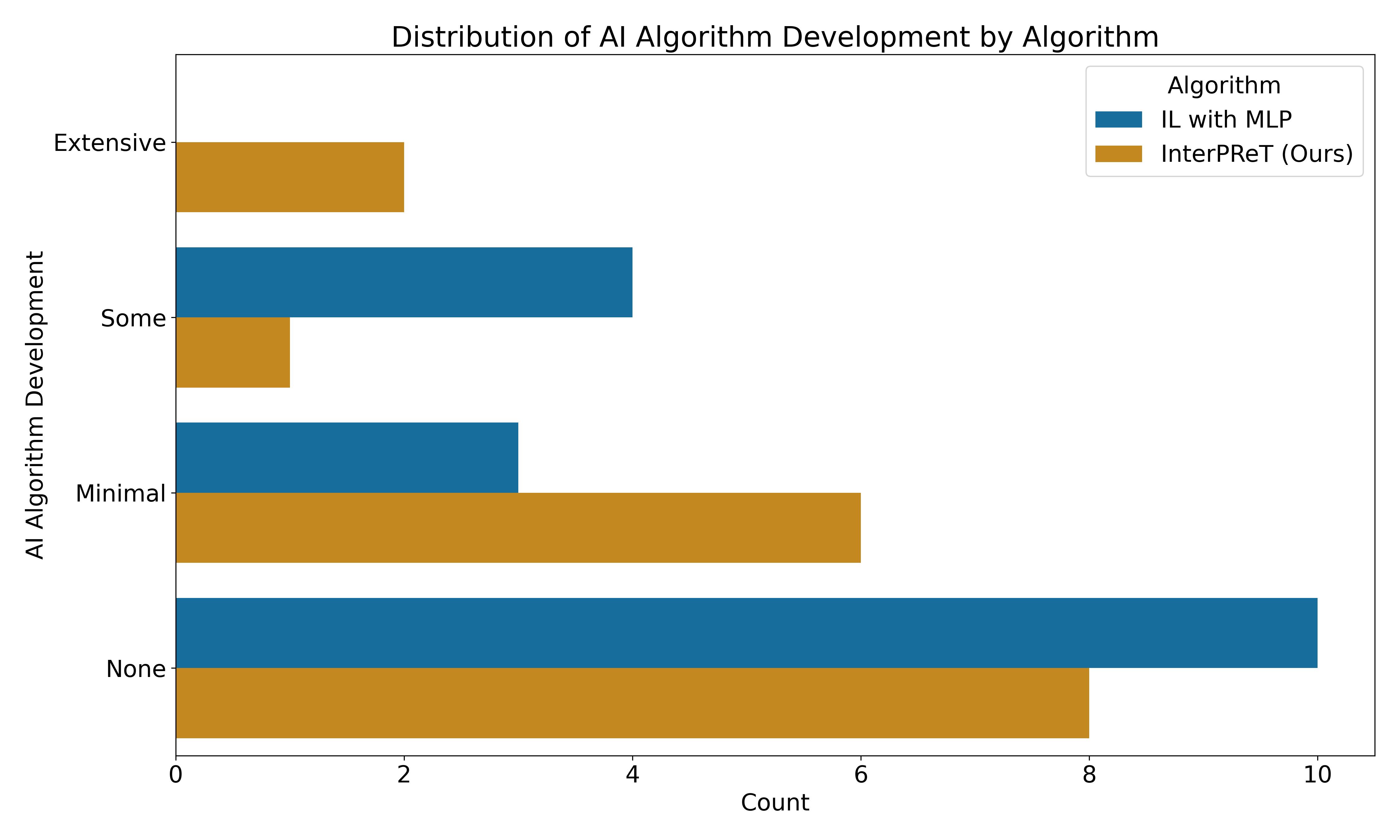}
    \caption{AI Development Experience Distribution}
    \label{fig:count-dev}
    \Description{
    A count plot that shows the distribution of AI development experience.
    Specifically, for the baseline condition, 4 participants reported some experience, 3 reported minimal, and 10 reported none.
    For the interpret condition, 2 participants reported extensive experience, 1 reported some, 6 reported minimal, 8 reported none.
    }
\end{figure}

\begin{figure}[h]
    \centering
    \includegraphics[width=0.7\linewidth]{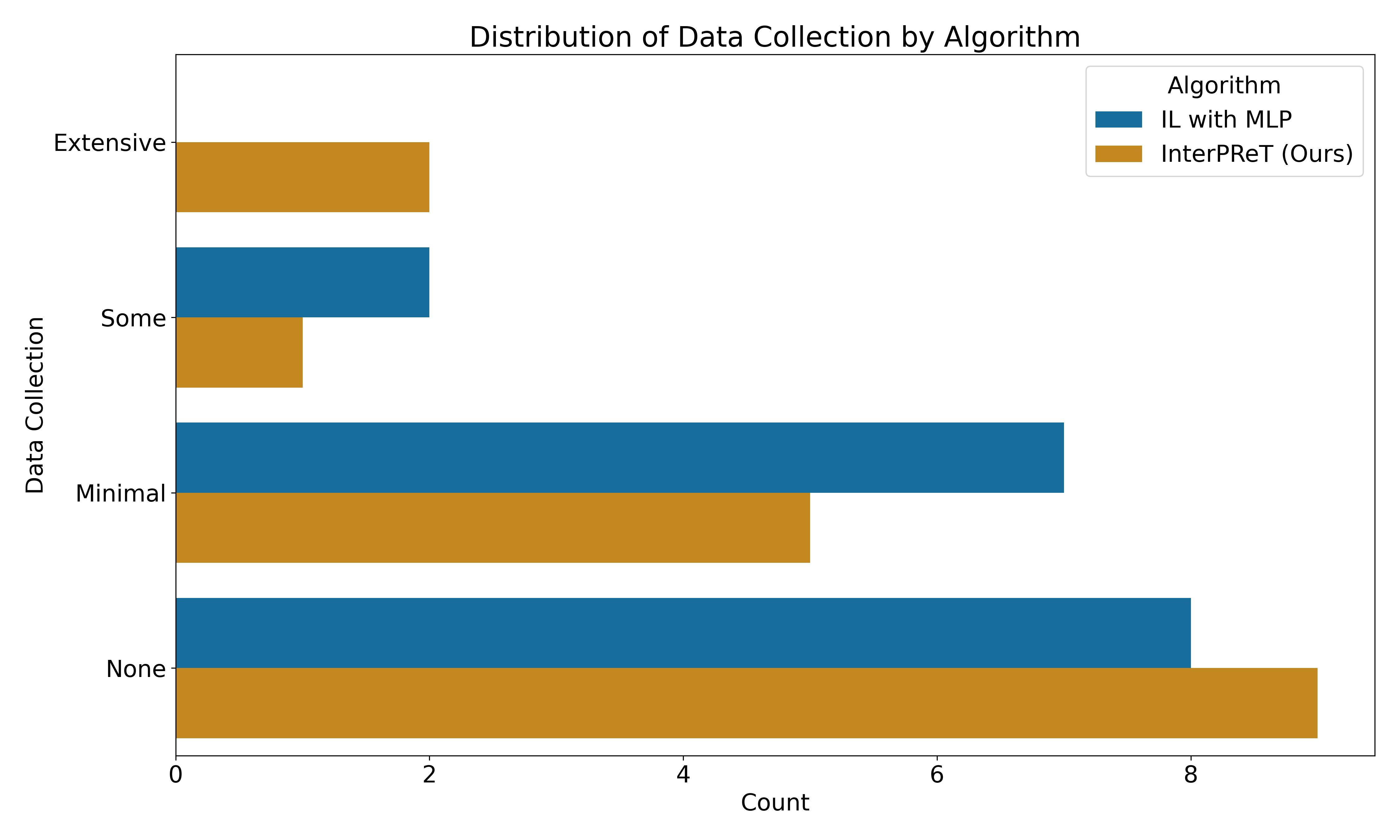}
    \caption{Data Collection Experience Distribution}
    \label{fig:count-data}
    \Description{
    A count plot that shows the distribution of data collection experience.
    Specifically, for the baseline condition, 2 participants reported some experience, 7 reported minimal, and 8 reported none.
    For the interpret condition, 2 participants reported extensive experience, 1 reported some, 5 reported minimal, 9 reported none.
    }
\end{figure}

There are five recruiting criteria for participating:
\begin{itemize}
    \item At least 18 years and no more than 64 years old
    \item Proficient in English
    \item Have normal or corrected-to-normal vision
    \item Can operate a handheld gamepad controller using both hands
    \item Do not have cognitive impairment that interferes with the use of a driving simulator.
\end{itemize}

Figures \ref{fig:count-age}, \ref{fig:count-gender}, \ref{fig:count-game}, \ref{fig:count-driving}, \ref{fig:count-dev}, \ref{fig:count-data} show the background distribution of the participants. 
Notably, $4$ of the participants in the baseline condition reported ``Video game every day or every other day" while none in the \interp{} condition had that extensive video game experience.
For other categories, the participants are mostly evenly distributed.

\newpage
\section{Learning Details}
For both conditions, we train with the Adam optimizer \cite{kingma2014adam} with a learning rate of $0.001$. Each batch has $512$ state and action pairs sampled from all demonstrations in the ``used for training" box. The MLP baseline has the shape $58 \times 80+80\times 3$ and is initialized randomly, while \interp{} has an initialization value acquired from GPT. All models are trained with $800$ batches regardless of the number of demonstrations provided. All inputs are normalized to $[-1, 1]$ and we use the mean squared error as the loss function. Empirically, the setup is enough for convergence. Typically \interp{} converges to a loss around $0.02$ while the baseline MLP can achieve a loss around $0.001$. This process takes about $3$ minutes on a modern CPU.

The total cost of the GPT query is less than $\$5$ for all participants. The wait time for each GPT query is around $3$ to $7$ minutes, depending on the length of the user instructions.

\end{document}